%% file: main.tex
\documentclass{article}

\usepackage{PRIMEarxiv}

\usepackage[utf8]{inputenc} 
\usepackage{cinzel}
\usepackage[T1]{fontenc}    
\usepackage{url}            
\usepackage{booktabs}       
\usepackage{amsfonts}       
\usepackage{nicefrac}       
\usepackage{microtype}      
\usepackage{lipsum}
\usepackage{fancyhdr}       
\usepackage{graphicx}       
\graphicspath{{media/}}     

\usepackage{xspace}
\usepackage{amsfonts}
\usepackage{amssymb}
\usepackage{pifont}
\usepackage{bm}
\usepackage{marvosym}

\usepackage{adjustbox}
\usepackage{multirow}

\usepackage{amsmath}
\usepackage{amssymb}
\usepackage{mathtools}
\usepackage{amsthm}
\usepackage{balance}
\usepackage{subcaption}

\newcommand\titlefont[1]{{\usefont{T1}{cinzeldecorativebold}{m}{n}#1}}

\newcommand{\onepeace}{ONE-PEACE\xspace}
\newcommand{\modeltitle}{\titlefont{ONE-PEACE}}
\newcommand{\modelname}{\titlefont{ONE-PEACE}\xspace}

\usepackage[dvipsnames]{xcolor}         
\usepackage{colortbl}
\usepackage{cuted}
\usepackage{makecell}

\definecolor{baselinecolor}{gray}{.9}

\definecolor{highlightcolor}{HTML}{eaf7eb}  

\newlength\savewidth\newcommand\shline{\noalign{\global\savewidth\arrayrulewidth
  \global\arrayrulewidth 1pt}\hline\noalign{\global\arrayrulewidth\savewidth}}
\newcommand{\tablestyle}[2]{\setlength{\tabcolsep}{#1}\renewcommand{\arraystretch}{#2}\centering\footnotesize}
\newcommand{\normaltablestyle}[2]{\setlength{\tabcolsep}{#1}\renewcommand{\arraystretch}{#2}\centering\normalsize}

\definecolor{citecolor}{RGB}{65,130,164}
\definecolor{linkcolor}{RGB}{166,64,54}
\definecolor{citecolor2}{HTML}{0071BC}
\usepackage[pagebackref=true,breaklinks=true,colorlinks,
citecolor=citecolor2,linkcolor=magenta,bookmarks=false]{hyperref}

\definecolor{dt}{gray}{0.7}  %

\newcolumntype{x}[1]{>{\centering\arraybackslash}p{#1pt}}
\newcolumntype{y}[1]{>{\raggedright\arraybackslash}p{#1pt}}
\newcolumntype{z}[1]{>{\raggedleft\arraybackslash}p{#1pt}}

\newcolumntype{*}{>{\global\let\currentrowstyle\relax}}
\newcolumntype{^}{>{\currentrowstyle}}
\newcommand{\rowstyle}[1]{\gdef\currentrowstyle{#1}#1\ignorespaces}

\newcommand{\tabincell}[2]{\begin{tabular}{@{}#1@{}}#2\end{tabular}}

\title{{\modeltitle}: Exploring one general Representation Model toward unlimited modalities}

\author{
   Peng Wang$^{1}$\thanks{Equal contribution. Work done during Shijie’s internship at DAMO Academy. $^\dag$Corresponding author} ,
   Shijie Wang$^{1,2*}$, Junyang Lin$^{1}$, Shuai Bai$^{1}$\\\\ 
   \textbf{Xiaohuan Zhou$^{1}$, Jingren Zhou$^{1}$, Xinggang Wang$^{2}$, Chang Zhou$^{1\dag}$
   } \\\\
   $^1$DAMO Academy, Alibaba Group\ \ \ $^2$Huazhong University of Science and Technology
}

\begin{document}
\maketitle

\begin{abstract}
In this work, we explore a scalable way for building a general representation model toward unlimited modalities.
We release \onepeace, a highly extensible model with 4B parameters that can seamlessly align and integrate representations across vision, audio, and language modalities.
The architecture of \onepeace comprises modality adapters, shared self-attention layers, and modality FFNs. 
This design allows for the easy extension of new modalities by adding adapters and FFNs, while also enabling multi-modal fusion through self-attention layers.
To pretrain \onepeace, we develop two modality-agnostic pretraining tasks, cross-modal aligning contrast and intra-modal denoising contrast, which align the semantic space of different modalities and capture fine-grained details within modalities concurrently.
With the scaling-friendly architecture and pretraining tasks, \onepeace has the potential to expand to unlimited modalities.
Without using any vision or language pretrained model for initialization, \onepeace achieves leading results on a wide range of uni-modal and multi-modal tasks, including image classification (ImageNet), semantic segmentation (ADE20K), audio-text retrieval (AudioCaps, Clotho), audio classification (ESC-50, FSD50K, VGGSound), audio question answering (AVQA), image-text retrieval (MSCOCO, Flickr30K), and visual grounding (RefCOCO/+/g).

\smallskip
\centering
Code is available at \url{https://github.com/OFA-Sys/ONE-PEACE}
\end{abstract}

\input{content/1_intro_new.tex}

\input{content/2_related_work.tex}
\input{content/3_method_new.tex}
\input{content/4_experiments.tex}

\input{content/5_conclusion.tex}

\section*{Acknowledgments}
We would like to thank Yang Zhang, Benjin Mei, Dongkun Li, Jin Wang, Wei Wang, and Yinghui Liu for their support to this project, and we would like to thank Yusong Wu for patiently answering our questions about the audio dataset. We would also thank the M6-Team for providing a supportive research environment.

\bibliographystyle{unsrt}  
\bibliography{main}

\newpage
\appendix
\input{content/6_appendix}

\end{document}

%% file: content/1_intro_new.tex
\section{Introduction}
\label{sec:intro}

Representation models have received considerable attention in computer vision~\cite{beit,mae,ibot,moco,mocov2,mocov3,simclr,dino,dinov2}, speech processing~\cite{wav2vec,wav2vec2,wavlm,hubert}, natural language processing~\cite{elmo,bert,roberta,electra,deberta}, etc. 
Learning from large amounts of data, representation models demonstrate strong generalization ability in a wide range of downstream tasks.
Furthermore, the explosive growth of large-scale language models (LLMs) has sparked an escalating appetite for representation models.
Until recently, representation models have shown their bedrock role to unleash LLMs to understand, perceive, and interact with other modalities (e.g., vision)~\cite{gpt4,kosmos,blip2,minigpt4,llava,mplug-owl,palme}.

Due to the distinct characteristics of different modalities, previous research mainly focuses on building uni-modal representation models with individual architectures and pretraining tasks.
Despite achieving excellent results, uni-modal representation models face difficulties in effectively utilizing multi-modal data such as image-text pairs and audio-text pairs, which makes them challenging to extend to multi-modal tasks.
With the development of unified architectures~\cite{transformer,vit,perceiver,perceiverio} and efficient pretraining tasks~\cite{bert,beit,mae,clip}, recent works have achieved promising results in vision-language learning~\cite{ofa,flamingo,coca,beit3,blip,pali} and audio-language learning~\cite{audioclip,wav2clip,clap,speecht5}. Nevertheless, there is still rare research on developing general models that can be applied to vision, audio, and language modalities.
\cite{beit3} utilize the Multiway Transformer to process both image and text modalities with a unified masked prediction task for pretraining. The masked prediction task requires a pretrained CLIP~\cite{clip} model to discretize image data, which limits the scalability to other modalities such as audio.
\cite{data2vec} proposes a general pretraining method that can be applied to vision, audio, and language modalities without the need for third-party models (e.g., CLIP), but it doesn't extend the method to multi-modal data.

In this paper, we explore a scalable way to build a general representation model toward unlimited modalities.
We advocate that a general representation model should meet the following conditions:
1. The model architecture must be flexible enough to accommodate various modalities and support multi-modal interaction. 
2. Pretraining tasks should not only extract information within each modality but also ensure alignment across modalities. 
3. Pretraining tasks should be general and straightforward, allowing them to be applied to different modalities.


Driven by these motivations, we propose \onepeace, a model with 4B parameters that can seamlessly align and integrate representations across vision, audio, and language modalities.
The architecture of \onepeace consists of multiple modality adapters and a modality fusion encoder.
Each modality is equipped with an adapter for converting the raw inputs into feature sequences.
The modality fusion encoder operates on feature sequences with Transformer architecture. Each Transformer block contains a shared self-attention layer and multiple modality Feed Forward Networks (FFNs).
The self-attention layer enables interaction between the multi-modal features through the attention mechanism, while the modality FFNs facilitate information extraction within modalities.
With the clear division of labor in this architecture, extending new modalities only requires the injection of adapters and FFNs.

To pretrain \onepeace, we design two modality-agnostic pretraining tasks.
The first one is cross-modal contrastive learning, it contains both vision-language contrastive learning and audio-language contrastive learning, which effectively align the semantic spaces of vision, audio, and language modalities.
The second one is intra-modal denoising contrastive learning, it can be viewed as a combination of masked prediction~\cite{bert,beit,mae,wav2vec} and contrastive learning~\cite{clip,simclr,moco}, where we perform contrastive loss between the fine-grained masked features and visible features, such as image patches, language tokens, or audio waveform features.
These tasks collaborate to enhance the model's fine-tuning performance while also maintaining cross-modal retrieval capability.
Furthermore, they are universal for all modalities, which obviates the need for modality-specific designs.
With the scaling-friendly model architecture and pretraining tasks, \onepeace has the potential to expand to unlimited modalities.

We conduct comprehensive experiments on different tasks across various modalities, including vision, audio, vision-language, and audio-language tasks.
Without using any vision or language pretrained model for initialization, \onepeace achieves leading results in both uni-modal and multi-modal tasks, including image classification ($89.8\%$ accuracy on ImageNet w/o privately labeled data), semantic segmentation ($63.0\%$ mIoU on ADE20K), audio-text retrieval (outperforming previous SOTAs on AudioCaps and Clotho by a large margin), audio classification ($91.8\%$ zero-shot accuracy on ESC-50, $69.7\%$ accuracy on FSD50K, $59.6\%$ accuracy on VGGSound w/o visual information), audio question answering ($86.2\%$ accuracy on AVQA w/o visual information), image-text retrieval ($84.1\%$ I2T R@1 on MSCOCO and $97.6\%$ I2T R@1 on Flickr30K w/o intermediate finetuning and ranking), and visual grounding ($89.26\%$/$83.23\%$/$89.27\%$ scores on RefCOCO/+/g test sets).



%% file: content/2_related_work.tex
\section{Related Work}
\label{sec:related-work}

\paragraph{Vision-Language Pretraining.}
Recent years have witnessed the rapid development of vision-language pretraining. 
Early approaches~\cite{vlbert,vilbert,visualbert,uniter,interbert,oscar,vinvl,unimo,vlt5,m6} relied heavily on region features extracted by object detectors, which is resource\&time-consuming. 
With the increasing popularity of Vision Transformer~\cite{vit}, numerous works use Transformer to jointly learn vision-language data and demonstrate superior performance in downstream tasks~\cite{vilt,albef,meter,vlmo,fiber,lemon,blip,blip2,mplug2}. 
To facilitate alignment between vision and language modalities, researchers propose various efficient pretraining tasks.
Among them, contrastive learning is one of the most representative methods that has been widely adopted in a lot of works~\cite{clip,align,lit,florence,simclr,moco,albef,chinese_clip}.
There also emerge some works explore the unified frameworks to handle vision-language tasks~\cite{ofa,simvlm,git,uni-perceiver,flava,coca,beit3,pali,metalm,unified_io}. \cite{ofa} adopts an encoder-decoder model to transform all vision-language tasks into generation tasks. 
\cite{coca} uses contrastive learning and text generation as the pretraining objectives, thus can be applied to image-text retrieval and vision-language generation tasks. 
\cite{beit3} employs the Multiway Transformer to process vision-language data, and discretizes images into image tokens through CLIP~\cite{clip} for joint learning with text tokens.

\paragraph{Audio-Language Pretraining.}

There is currently a significant amount of research being conducted in audio-language pretraining.
One category of these works focuses on speech-text joint pretraining. For instance, some studies propose to train a unified encoder for speech and text, which utilizes a large amount of unlabeled speech and text with masked prediction tasks and paired speech-text data to learn alignment~\cite{slam,mslam,maestro,speechlm}. 
There are also some works proposed to jointly pretrain speech and text under the encoder-decoder framework~\cite{speecht5,stpt,speechut,mmspeech}, which can be well applied to generation tasks, such as speech recognition and synthesis. 
Another category introduces cross-modal contrastive learning to audio-language pretraining~\cite{audioclip,clap,wav2clip,laion_clap}. \cite{clap} uses CNN14~\cite{panns} and BERT~\cite{bert} to extract audio and text information respectively, and conducts contrastive learning on environmental sound data. 
\cite{laion_clap} further introduces more environmental sound data and trained with HTSAT~\cite{htsat} and RoBERTa~\cite{roberta}. 
It achieves state-of-the-art results in audio downstream tasks such as audio classification and audio-text retrieval.


\paragraph{Vision-Audio-Language Pretraining.}
Recently researchers have begun exploring joint learning of vision, audio, and language modalities. 
\cite{vatt} employs a unified Transformer model to jointly learn video, text, and audio through cross-modal contrastive learning. 
\cite{icode,vatlm} utilize external models (e.g., VQ-VAE~\cite{vqvae} and AV-HuBERT~\cite{av_hubert}) to discretize the video and audio data, and train the models with masked prediction objectives.
~\cite{data2vec} proposes a general self-supervised learning method that does not rely on external models.
It successfully applies the method to vision, language, and audio modalities, but has not extended to multi-modal data.


Compared to previous works, \onepeace has a flexible architecture that is compatible with multiple modalities. Furthermore, the pretraining tasks of \onepeace are universally applicable without external models
Therefore, \onepeace can be easily extended to various modalities.

%% file: content/3_method_new.tex
\section{Method}
\label{sec:method}

\begin{figure*}[t]
\vskip 0.2in
    \centering
    \includegraphics[width=0.85\linewidth]{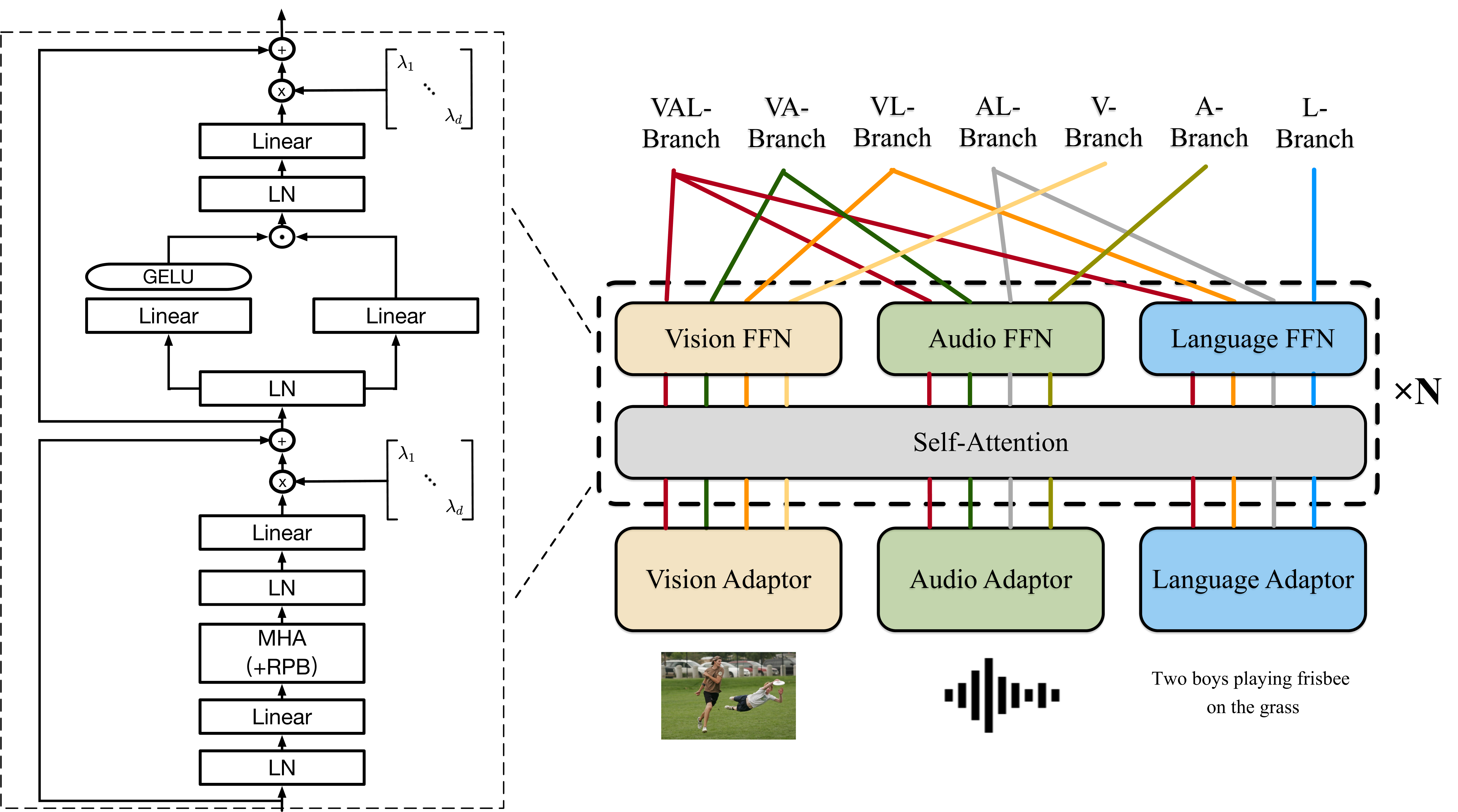}
    \caption{\textbf{The architecture of \onepeace}. 
    It consists of three modality adapters and a modality fusion encoder.
    \onepeace can be disassembled into different branches to handle different tasks.
    For example, the vision adapter, self-attention layers, and vision FFNs can be combined into V-Branch to handle vision tasks.}
    \label{fig:model}
\end{figure*}

\subsection{Architecture}
The model architecture of \onepeace consists of three modality adapters and a modality fusion encoder.
The overall architecture is shown in Figure~\ref{fig:model}.

\paragraph{Modality Adapters.}
We design modality adapters to convert different raw signals into unified features.
Note that these adapters do not interact with each other, which affords us the flexibility to choose appropriate networks for them, such as Transformers~\cite{transformer,vit,swin}, CNNs~\cite{mnist,resnet}, RNNs~\cite{rnn,gru}, etc.
We design three lightweight modality adapters for \onepeace:

\begin{itemize}
    \item \textbf{Vision Adapter (V-Adapter).} Given an image, we use a hierarchical MLP (hMLP) stem~\cite{Touvron2022ThreeTE} to patchify the image by gradually increasing the patch size to $16 \times 16$. There is no interaction between different patches. Then the image patches are flattened into a sequence and prepended with a vision class embedding. By adding the absolute positional embeddings into the image embeddings, the image representation is $E^V=\langle \bm{e}^V_{cls}, \bm{e}^V_{1}, \bm{e}^V_{2}, ..., \bm{e}^V_{M} \rangle$, where $M$ denotes the total number of image patches.
    \item \textbf{Audio Adapter (A-Adapter).} Given an audio, we set the sample rate to 16kHz and normalize the raw audio waveform to zero mean and unit variance. Then the normalized waveform is processed by a convolutional feature extractor~\cite{wav2vec2} to get the audio embeddings. Instead of using the absolute positional embeddings, we use a convolution layer to extract relative position information and add it to the audio embeddings~\cite{Mohamed2019TransformersWC}. With a prepended audio class embedding, we obtain the audio representation $E^A=\langle \bm{e}^A_{cls}, \bm{e}^A_{1}, \bm{e}^A_{2}, ..., \bm{e}^A_{N} \rangle$, where $N$ denotes the length of the audio representation.
    \item \textbf{Language Adapter (L-Adapter).} Given a text, we first apply byte-pair encoding (BPE) \cite{bpe} to transform it to a subword sequence. Two special tokens $\left[{\rm CLS}\right]$ and $\left[{\rm EOS}\right]$ are inserted at the beginning and end of the sentence to indicate its start and end. Then an embedding layer is used to embed the subword sequence to the text embeddings. After summing the text embeddings with absolute positional embeddings, we obtain the text representation $E^L=\langle \bm{e}^L_{cls}, \bm{e}^L_{1}, \bm{e}^L_{2}, ..., \bm{e}^L_{K}, \bm{e}^L_{eos} \rangle$, where $K$ denotes the text sequence length.
\end{itemize}

\paragraph{Modality Fusion Encoder.} 
Following previous works~\cite{simvlm,ofa,coca,beit3,ofasys}, the modality fusion encoder is based on the Transformer architecture~\cite{transformer}. 
We set up a shared self-attention layer and three modality feed-forward networks (FFNs) in each Transformer block.
The shared self-attention layer enables the interaction between different modalities through the attention mechanism.
The three modality FFNs (V-FFN, A-FFN, and L-FFN) can further extract information within their respective modalities.
To stabilize training and enhance model performance, we make the following improvements:

\begin{itemize}
    \item \textbf{Sub-LayerNorm.} We incorporate Sub-LayerNorm~\cite{magneto} into each Transformer block to enhance training stability. Specifically, We insert layer normalization before the input projection and output projection of each self-attention layer and FFN layer. In our preliminary experiments, we find that Sub-LayerNorm can achieve better performance compared to the Pre-LayerNorm~\cite{gpt3}.
    \item \textbf{GeGLU Activation Function.} To further improve performance, we replace the activation function in FFN with GeGLU~\cite{glu} activation function. The intermediate dimension of FFN is set to $4$ times the embedding dimension, which is consistent with the practice of PaLM~\cite{palm}.
    \item \textbf{Relative Position Bias (RPB).} For positional information, we introduce 1D relative position bias~\cite{T5} for text and audio, and 2D relative position bias for image~\cite{coatnet}. At the pretraining stage, the relative position bias of different self-attention layers is shared. At the fine-tuning stage, we decouple the relative position bias of each self-attention layer and let them inherit the weights of the pretrained relative bias.
    \item \textbf{LayerScale.} We use LayerScale~\cite{cait} to dynamically adjust the output of each residual block. Specifically, before adding to the residual, we multiply the output of each layer (e.g., self-attention layer and FFN) by a learnable diagonal matrix, whose values will be initialized to $1e-6$. In our preliminary experiments, LayerScale is beneficial for stabilizing training and improving performance.
\end{itemize}

This "sharing-separated" architecture enables \onepeace to disassemble into different branches that handle tasks for various modalities.
For example, the vision adapter, self-attention layer, and vision FFNs can be combined into the vision branch (V-Branch) to process vision tasks.
Similarly, we named other branches as audio branch (A-Branch), language branch (L-Branch), vision-audio branch (VA-Branch), vision-language branch (VL-Branch), audio-language branch (AL-Branch), and vision-audio-language branch (VAL-Branch).

\begin{figure*}[t]
\vskip 0.2in
    \centering
    \includegraphics[width=0.8\linewidth]{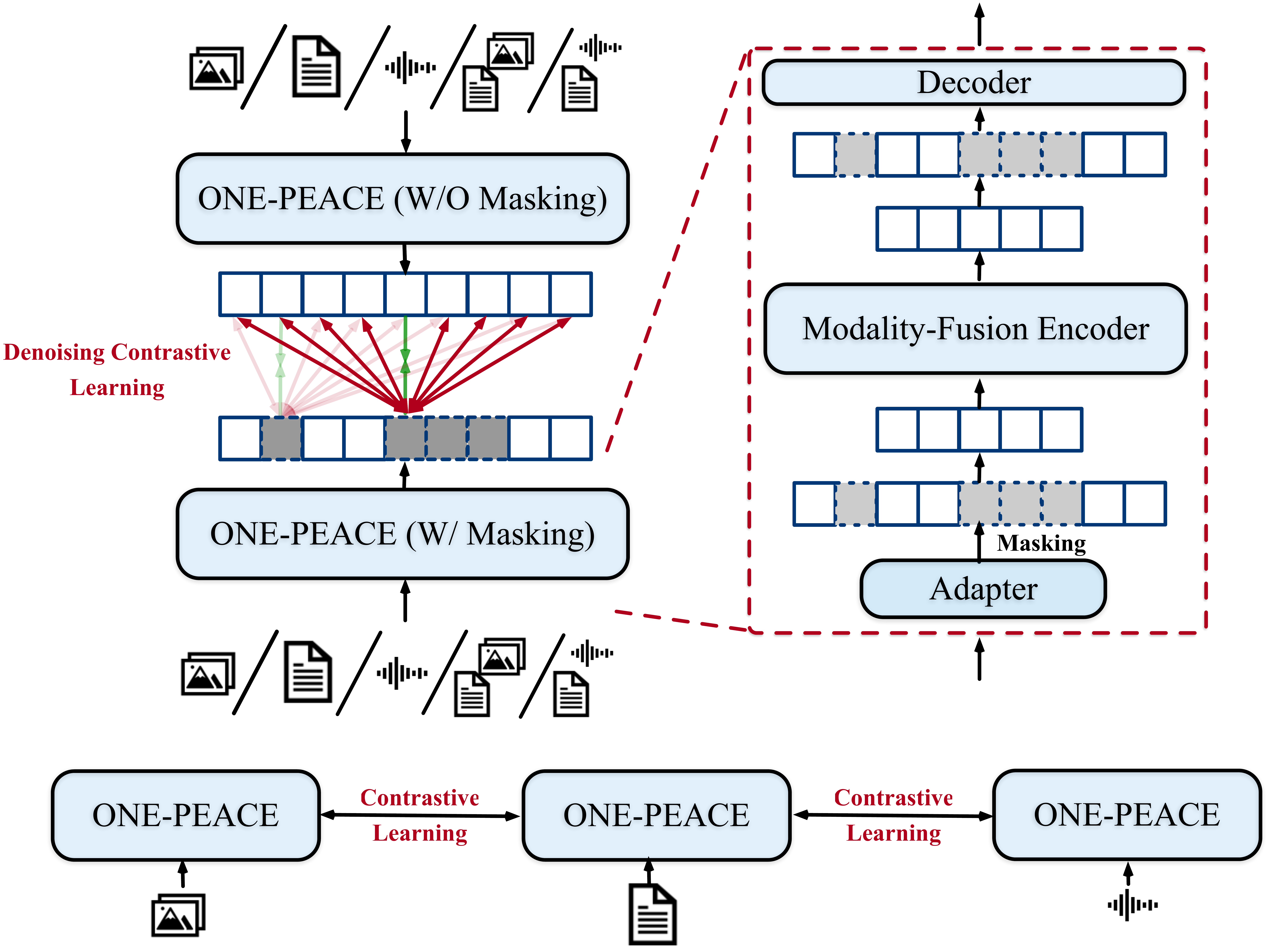}
    \caption{\textbf{The pretraining tasks of \onepeace.} Intra-modal denoising contrastive learning encourages the features of the masked units (e.g., image patches or text tokens) close to the positive units (indicated by the green lines) and get away from the negative units (indicated by the red lines). Note that we compute the cross-modal contrastive loss by gathering negative features from all GPU devices, while the denoising contrastive loss is computed on the local batch.}
    \label{fig:loss}
\end{figure*}

\subsection{Tasks}
The pretraining tasks of \onepeace include cross-modal contrastive learning and intra-modal denoising contrastive learning. 
Cross-modal contrastive learning endows the model with cross-modal retrieval capability, while intra-modal denoising contrastive learning enables the model to achieve superior fine-tuning performance in downstream tasks. 
An illustration of the pretraining tasks is shown in Figure~\ref{fig:loss}.

\paragraph{Cross-Modal Contrastive Learning.}

Cross-modal contrastive learning is a widely-used pretraining task that effectively aligns the semantic spaces of different modalities. 
The key idea of this method is to maximize the similarity of related sample pairs across different modalities while minimizing the similarity of unrelated sample pairs.
Given a sample pair $(S^1, S^2)$ of arbitrary modalities (e.g., image-text pair or audio-text pair), we extract their features using the corresponding branches of \onepeace.
The outputs of the special tokens (e.g., vision class token or language class token) are regarded as global representations. Followed by a linear projection and normalization, we obtain the final representations $\bm{s}^1$ and $\bm{s}^2$. The loss function is shown below:

\begin{equation}
\mathcal{L}_{CL} = -\frac{1}{2N}\sum_{i=1}^{N}({\rm log}\frac{{\rm exp}(\bm{s}_i^{1}\bm{s}_i^{2}/\sigma)}{\sum_{j=1}^{N}{\rm exp}(\bm{s}_i^1\bm{s}_j^2/\sigma)} + {\rm log}\frac{{\rm exp}(\bm{s}_i^{1}\bm{s}_i^{2}/\sigma)}{\sum_{j=1}^{N}{\rm exp}(\bm{s}_j^1\bm{s}_i^2/\sigma)}),
\end{equation}

where $N$ is the batch size, $i,j$ are indexes within the batch, and $\sigma$ is a learnable temperature parameter (initialized to $0.07$). Following previous works~\cite{clip,align}, the cross-modal contrastive loss is computed by gathering negative features from all GPU devices.
We apply cross-modal contrastive learning to image-text pairs and audio-text pairs, denoted by $\mathcal{L}_{CL-VL}$ and $\mathcal{L}_{CL-AL}$ respectively.

\paragraph{Intra-Modal Denoising Contrastive Learning.}


Cross-modal contrastive learning mainly focuses on aligning features between different modalities. 
However, it lacks emphasis on the learning of fine-grained details within modalities, leading to suboptimal performance in downstream tasks~\cite{fd-clip}. 
To address this issue, we further introduce intra-modal denoising contrastive learning to train \onepeace\footnote{Intra-modal denoising contrastive learning is similar to \cite{conmim}, but extends to more modalities.}.
Intra-modal denoising contrastive learning can be viewed as a combination of masked prediction and contrastive learning, where we perform contrastive loss between the fine-grained masked features and visible features, such as image patches, text tokens, or audio waveform features.

Given a sample of arbitrary modalities, we first encode it into an embedding sequence through the corresponding modality adapter.
Then, we randomly mask some units (e.g., text tokens or image patches) within the sequence.
Following~\cite{mae}, we only input the unmasked units to the modality fusion encoder to reduce computation costs and save memory.
The encoded unmasked features are concatenated with the learnable mask tokens and fed to a lightweight Transformer decoder, which generates the masked features.
We also use the \onepeace model to encode the raw input sample into target features without masking.
Finally, we perform the contrastive loss between the masked features and target features, the loss function is shown below:

\begin{equation}
\label{eq:mcl}
\mathcal{L}_{DCL} = -\frac{1}{N\hat{N}}\sum_{i=1}^{N}\sum_{j=1}^{\hat{N}}{\rm log}\frac{{\rm exp}(\bm{\hat{h}}_{ij} \cdot {\rm sg}(\bm{h}_{ij})/\tau)}{\sum_{m=1}^{N}\sum_{n=1}^{N}{\rm exp}(\bm{\hat{h}}_{ij}  \cdot {\rm sg}(\bm{h}_{mn})/\tau)},
\end{equation}

Where $\bm{\hat{h}}_{ij}$ is the representation of the masked unit, $\bm{h}_{ij}$ is the representation of the target unit, $\text{sg}(\cdot)$ is the stop gradient operation. $\hat{N}$ is the number of masked units within a sample, $N$ is the number of whole units within a sample. $\tau$ is a constant temperature value, we set it to $0.4$. 
This loss not only encourages the masked units close to the positive units but also gets away from the negative units.
As a result, each unit acquires a unique semantic meaning, which makes the model better transfer to downstream tasks~\cite{fd-clip}.

We apply intra-modal denoising contrastive learning to $5$ types of data: image, audio, text, image-text pairs, and audio-text pairs.
For image, we randomly mask $75\%$ patches, and the loss function used for this type of data is denoted by $\mathcal{L}_{DCL-V}$.
For audio, we sample $p=0.11$ of all time-steps to be starting indices and mask the subsequent $5$ time-steps, and the loss function is denoted by $\mathcal{L}_{DCL-A}$.
For text, we randomly mask $15\%$ tokens of the text sequence, and the loss function is denoted by $\mathcal{L}_{DCL-L}$.
For image-text pairs, we randomly mask $68.75\%$ patches of the image and $40\%$ tokens of the text. The unmasked patches and tokens are concatenated together and encoded as masked features.
The original image patches and text tokens are also concatenated together and encoded as target features.
We then perform contrastive loss on the image patches and text tokens respectively, the average of these two
losses is denoted by $\mathcal{L}_{DCL-VL}$.
For audio-text pairs, we randomly mask $45\%$ time-steps of the audio waveform and $40\%$ tokens of the text. 
The loss is similar to the above one, we denote it by $\mathcal{L}_{DCL-AL}$.

\subsection{Training}
\label{sec:two_stage_pretraining}
The overall pretraining process of \onepeace is divided into two stages: vision-language pretraining and audio-language pretraining.
At the vision-language pretraining stage, the model trains on image-text pairs and only updates parameters that are relevant to vision and language modalities.
For each image-text pair, we not only utilize them to calculate $\mathcal{L}_{CL-VL}$ and $\mathcal{L}_{DCL-VL}$, but also separately using the image and text to calculate $\mathcal{L}_{DCL-V}$ and $\mathcal{L}_{DCL-L}$ respectively.
The loss function at this stage is shown below:

\begin{equation}
\mathcal{L}_{VL} = \mathcal{L}_{CL-VL} + 1.0*\mathcal{L}_{DCL-V} + 0.5*\mathcal{L}_{DCL-L} + 1.0*\mathcal{L}_{DCL-VL}
\end{equation}

At the audio-language pretraining stage, the model trains on audio-text pairs, and we only update A-Adapter, A-FFNs, and other audio-related parameters. 
The remaining parameters including self-attention layers are totally frozen.
Despite not training on image-audio pairs, the semantic space between vision and audio is still aligned by using language as the anchor.
The loss function at the audio-language pretraining stage is shown below:

\begin{equation}
\mathcal{L}_{AL} = \mathcal{L}_{CL-AL} + 1.0*\mathcal{L}_{DCL-A} + 1.0*\mathcal{L}_{DCL-AL}
\end{equation}


\section{Pretraining Details}
\input{table/model_size}

\paragraph{Pretraining Datasets.}
The pretraining datasets of \onepeace are divided into two parts: image-text pairs and audio-text pairs. 
For image-text pairs, we use LAION-2B~\cite{laion5b}, a dataset obtained by web crawling.
For audio-text pairs, we collect a large amount of open-source environmental sound datasets.
To ensure reproducibility, all pretraining datasets are publicly available.
We provide more details about the pretraining datasets in Appendix~\ref{app:audio_text_data_details}.



\paragraph{Pretraining Settings.}
\input{table/imagenet_result}
\onepeace is a giant-size model with $4$B parameters. We list the detailed hyper-parameters in Table~\ref{tb:model_configuration}. 
During pretraining, we introduce a lightweight Transformer decoder to recover the masked units from the visible units. The decoder is similar to the modality-fusion encoder, each block of it also consists of a shared self-attention layer and three modality FFNs.
It has $2$ layers with $768$ hidden size, $2048$ intermediate size, and $12$ attention heads.
The model weights of \onepeace are randomly initialized at the beginning, except for the audio feature extractor of A-adapter, for which we use the weights of WavLM's feature extractor~\cite{wavlm} for initialization. 
We find that incorporating WavLM's feature extractor significantly improves the model performance.
More details about the pretraining settings are provided in Appendix~\ref{app:pretraining_hyperparameters}.

\paragraph{Training Acceleration.}
We introduce several model acceleration and memory optimization techniques to accelerate the training.
Firstly, we use the memory-efficient attention technique~\cite{memory_efficient_attn,flash_attn} implemented in the xformers library\footnote{\url{https://github.com/facebookresearch/xformers}} to improve training speed.
Secondly, we use the gradient checkpointing technique~\cite{checkpointing} to save memory, which allows us to train the model with a larger batch size.
Furthermore, we replace the layer normalization with Fused LayerNorm implemented in the Flash Attention library\footnote{\url{https://github.com/HazyResearch/flash-attention}}, and leverage nvFuser\footnote{\url{https://pytorch.org/blog/introducing-nvfuser-a-deep-learning-compiler-for-pytorch}} to fuse the operations of dropout, LayerScale, stochastic depth, and residual summing, which can bring additional speed improvements.
To improve the training speed and prevent gradient overflow issues, we adopt Bfloat16 precision to train \onepeace.

%% file: table/model_size.tex
\begin{table*}[t]
\centering
\tablestyle{4pt}{1.2}
\begin{tabular}{cccccccccccc}
  \multirow{2}{*}{\#Layers}
  & \multirow{2}{*}{\bf \tabincell{c}{\scriptsize Hidden \\ \scriptsize Size}}
  & \multirow{2}{*}{\bf \tabincell{c}{\scriptsize Intermediate \\ \scriptsize Size}}
  & \multirow{2}{*}{\bf \tabincell{c}{\scriptsize Attention \\ \scriptsize Size}}
  & \multicolumn{8}{c}{\textbf{\#Parameters}}
  \\
  \cmidrule(lr){5-12}
  &
  &
  &
  & \textbf{\scriptsize V-Adapter}
  & \textbf{\scriptsize A-Adapter}
  & \textbf{\scriptsize L-Adapter}
  & \textbf{\scriptsize V-FFN}
  & \textbf{\scriptsize A-FFN}
  & \textbf{\scriptsize L-FFN}
  & \textbf{\scriptsize Shared Attention}
  & \textbf{\scriptsize Total}
  \\
  \shline
  40
  & 1536
  & 6144
  & 24
  & 3.4M
  & 19M
  & 78M
  & 1.15B
  & 1.15B
  & 1.15B
  & 378M
  & 4B
  \\
\end{tabular}
\caption{\textbf{Detailed hyperparameters of \onepeace model configuration.}}
\label{tb:model_configuration}
\end{table*}

%% file: table/imagenet_result.tex
\begin{table*}[t]
\centering
\normaltablestyle{6pt}{1.2}
\begin{tabular}{*l| ^c ^c ^c| ^c}
Method                                 & Enc. \#Params & Patch Size & Image size & Top-1 acc \\
\shline
FD-SwinV2-G~\cite{wei2022contrastive}  & 3.0B  & $16\times16$ & 336$^2$    & 89.4 \\
InternImage~\cite{wang2022internimage} & 1.08B & $16\times16$ & 640$^2$    & 89.2 \\
BEiT-3~\cite{beit3}                    & 1.01B & $14\times14$ & 336$^2$    & 89.6 \\
EVA~\cite{eva}                         & 1.01B & $14\times14$ & 560$^2$    & 89.7 \\
\hline
\modelname                             & 1.52B & $16\times16$ & 384$^2$    & 89.6 \\
\modelname                             & 1.52B & $16\times16$ & 512$^2$    & \textbf{89.8}
\\
\hline
\multicolumn{5}{l}{\rowstyle{\color{dt}}\textit{methods using extra privately collected data:}}
\\
\hline
\rowstyle{\color{dt}}RevCol-H~\cite{cai2022reversible}      & \color{dt}2.16B & - & 640$^2$    & 90.0 \\
\rowstyle{\color{dt}}ViT-G~\cite{modelsoups}      & \color{dt}1.84B & $14\times14$ & 518$^2$    & 90.5 \\
\rowstyle{\color{dt}}Model Soups~\cite{modelsoups}      & \color{dt}1.84B & $14\times14$ & 500$^2$    & 90.9 \\
\rowstyle{\color{dt}}CoCa~\cite{coca}      & \color{dt}1.01B & $18\times18$ & 576$^2$    & 91.0 \\
\end{tabular}
\caption{\textbf{System-level comparisons of image classification with the leading results on ImageNet-1k.} RevCol~\cite{cai2022reversible} is pretrained on a privately collected 168-million-image dataset. ViT-G~\cite{modelsoups} and Model Soups~\cite{modelsoups} are pretrained on JFT-3B~\cite{jft} with supervision. CoCa~\cite{coca} uses JFT-3B~\cite{jft} and ALIGN~\cite{align} datasets for pretraining. \onepeace achieves state-of-the-art results using the publicly available dataset with less token length.}
\label{tb:inet1k}
\end{table*}

%% file: content/4_experiments.tex
\section{Experiments}
\label{sec:experiments}

\subsection{Results on Vision Tasks}
\input{table/v_result}

We transfer \onepeace to various mainstream vision benchmarks, including image classification, semantic segmentation, object detection, instance segmentation and video action recognition. 
We provide the implementation details in Appendix~\ref{app:vision_details}.

\paragraph{Image Classification.} In our experiments, we assess the image classification transfer performance of \onepeace using the ImageNet-1K~\cite{inet1k} dataset, encompassing 1.28 million training images and 50,000 validation images distributed across 1,000 distinct categories. We also use intermediate fine-tuning on ImageNet-21k~\cite{imagenet}. As demonstrated in Table~\ref{tb:inet1k}, \onepeace obtains \textbf{89.8} top-1 accuracy on ImageNet with less token length $(image\_size / patch\_size)^2$. 
Note that FD-SwinV2-G, BEiT-3, and EVA all rely on the assistance of an external CLIP model for pretraining, while \onepeace is trained from scratch without the help of external models. Even so, \onepeace is able to achieve better results, which demonstrates its strong transferability.

\paragraph{Semantic Segmentation.} We experiment on ADE20k~\cite{Zhou2016SemanticUO} using ViT-Adapter~\cite{chen2022vitadapter} for task adaptation and Mask2Former~\cite{mask2former} as the segmentation head. Following common practice, we first fine-tune the segmentation head on coco-stuff~\cite{cocostuff} then fine-tune on ADE20k. As demonstrated in Table~\ref{tb:ade20k}, \onepeace establishes a new state-of-the-art, achieving a mean Intersection over Union (mIoU) of \textbf{63.0}. This result indicates that \onepeace exhibits exceptional transferring performance in the domain of dense prediction tasks.

\paragraph{Object Detection and Instance Segmentation.} We perform fine-tuning experiments on the COCO 2017~\cite{mscoco} dataset. For the backbone, we employ the \onepeace backbone and use the ViTDet~\cite{vitdet} with Cascade Mask-RCNN architecture, which incorporates a straightforward feature pyramid and window attention for addressing object detection and instance segmentation tasks. The model is fine-tuned on the COCO dataset. Soft-NMS~\cite{softnms} is used during the inference stage. As illustrated in Table~\ref{tb:coco}, the instance-level transfer capabilities of \onepeace exhibit a performance that is on par with the current state-of-the-art methods. 

\paragraph{Video Action Recognition.} We benchmark \onepeace on Kinetics 400~\cite{k400} dataset for video action recognition. Following AIM~\cite{aim}, we keep the whole model frozen and add several MLP adapters in each transformer layer. We use I3D~\cite{k400} head as the classification layer. As demonstrated in Table~\ref{tb:k400}, without fine-tuning the full encoder, ~\onepeace could achieve \textbf{88.1} top-1 accuracy, even outperforming CoCa which is pre-trained on privately collected data, and ViT-22B with 14x more parameters.

\subsection{Results on Audio(-Language) Tasks}
\input{table/audio_result}
We evaluate \onepeace on various audio and audio-language tasks, including audio-text retrieval, audio classification, and audio question answering (AQA).
The implementation details are provided in Appendix~\ref{app:audio_details}.

\paragraph{Audio-Text Retrieval.}
Table~\ref{tb:audio_text_retrieval} presents the performance of \onepeace and baseline models in the audio-text retrieval task.
As a general representation model, \onepeace achieves SOTA results on both AudioCaps~\cite{audiocaps} and Clotho~\cite{clotho} datasets, outperforming the previous audio representation model by a large margin.
On AudioCaps, \onepeace achieves 21.1\% improvement on R@1 in text-to-audio retrieval and 11.4\% improvement on R@1 in audio-to-text retrieval.
On Clotho, \onepeace achieves 23.1\% improvement on R@1 in text-to-audio retrieval and 5.4\% on R@1 in audio-to-text retrieval.

\paragraph{Audio Classification \& Audio Question Answering.}
Table~\ref{tb:audio_result} present the results of \onepeace and baseline models in the audio classification and audio question answering (AQA) tasks.
On ESC-50, \onepeace achieves $91.8$ zero-shot accuracy, outperforming LAION-CLAP by $0.8$.
On FSD50K, \onepeace significantly outperforms the previous SOTA by $4.1$.
For the VGGSound dataset, which consists of both visual and audio information, we only utilized the audio information and disregarded the visual information. 
With this setting, \onepeace achieves $59.6$ score, surpassing the previous SOTA by $0.1$.
In the audio question answering task, \onepeace outperforms the previous SOTA by $2.7$.
These results demonstrate the superior ability of \onepeace on audio-related tasks.

\subsection{Results on Vision-Language Tasks}
\input{table/image_text_result}

\input{table/vl_result}

We conduct experiments on various vision-language tasks, including image-text retrieval, visual grounding, visual question answering, and visual reasoning. the implementation details are provided in Appendix~\ref{app:vision_language_details}.

\paragraph{Image-Text Retrieval.}
Table~\ref{tb:vl-zeroshot} presents the performance of \onepeace and baseline models on the image-text retrieval task. 
Under the fine-tuning setting, \onepeace achieves the best performance in both MSCOCO and Flickr30K test sets. 
This indicates that after combining both cross-modal contrastive learning and intra-modal denoising contrastive learning, \onepeace can effectively transfer to downstream retrieval task.
Under the zero-shot setting, \onepeace can achieve better or competitive performance compared to previous dual-encoder models like CLIP and Florence.
Notice that the results of \onepeace are inferior to CoCa, which might be because \onepeace only trained on $6.4$ billion image-text pairs while CoCa trained on up to $32$ billion image-text pairs.

\begin{figure*}[t]
    \centering
    \includegraphics[width=\linewidth]{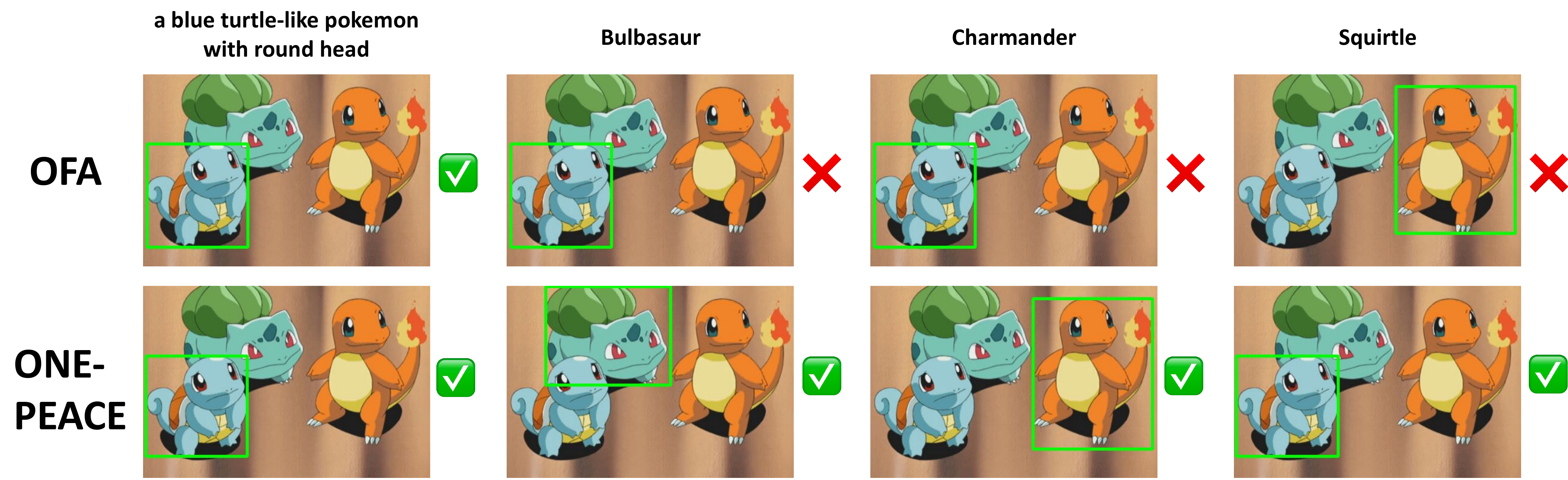}
    \caption{\textbf{Visualization of out-of-domain data in visual grounding task.} When given a specific description, both OFA and ONE-PEACE can give the correct result. However, OFA is unable to locate the correct region if we directly provide the name of the Pokémon, while ONE-PEACE can give a correct answer.}
    \label{fig:grounding_visual}
    \vspace{+1em}
\end{figure*}

\paragraph{Visual Grounding.}
To evaluate the capability of visual grounding, we conduct experiments on RefCOCO, RefCOCO+, and RefCOCOg datasets~\cite{refcoco,refcocog}. 
Table~\ref{tb:refcoco} presents the results of \onepeace and baseline models. 
It is worth noting that previous SOTA OFA use additional visual grounding datasets for training (i.e., Visual Genome~\cite{vg}). 
Without introducing additional visual grounding datasets, \onepeace still achieves new SOTA results on the $3$ datasets. 
We also compared the visual grounding ability of \onepeace and OFA on an out-of-domain Pokémon picture.\footnote{We use the Hugging Face spaces demo of OFA: \url{https://huggingface.co/spaces/OFA-Sys/OFA-Visual\_Grounding}} 
As shown in Figure~\ref{fig:grounding_visual}, given a specific description of a Pokémon, both \onepeace and OFA can obtain the correct result. 
However, when we directly provide the name of the Pokémon, OFA fails to obtain the correct result while ONE-PEACE can give a correct answer.

\paragraph{Vision-Language Understanding.}
\input{table/vision_language_understanding}
Table~\ref{tb:vqa-nlvr-ve} presents the results of \onepeace and baselines on two popular multimodal understanding tasks: visual question answering (VQA~\cite{vqa}) and visual reasoning (NLVR-2~\cite{nlvr2}). 
For the VQA task, \onepeace achieves a score of $82.6$ on the test-dev set and $82.5$ on the test-std set, outperforming previous strong baselines like CoCa and BLIP-2.
For the NLVR2 task, \onepeace surpasses CoCa with gains of 1.7 and 1.3 on the dev set and test-P set respectively.
Notice that our results on both tasks are lower than BEiT-3. This may be attributed to two reasons: Firstly, BEiT-3 is pretrained on in-domain datasets such as MSCOCO~\cite{mscoco} and Visual Genome~\cite{vg}, which usually results in better downstream finetuning effects. 
Secondly, BEiT-3 incorporates pure text data for pretraining, which improves its language understanding ability and consequently enhances its multimodal understanding ability.
In addition, OFA and BLIP-2 have shown that combined with language pretrained models can improve performance on multimodal understanding tasks. 
Therefore, we will explore the combination of \onepeace and language pretrained models in the future.

\subsection{Ablation Study}
\label{sec:ablation}
\input{table/ablation}

For the following ablation experiments, we utilize VIT-B/16 as the model backbone. The model is trained for 20 epochs with a batch size of 4096. We randomly selected 20 million image-text pairs from Laion-2B as the pretraining dataset.

\paragraph{Ablation on Model Structures.}
We first conduct ablation experiments to investigate the effects of sharing or separating different modules.
As shown in Table~\ref{tb:structure_ablation}, sharing both self-attention layers and FFN layers yields better results compared to not sharing. This suggests that utilizing a single Transformer can effectively align the semantic space of vision and language.
Furthermore, it is more beneficial to separate the FFN layer instead of sharing it. This implies that separating the FFN layer enhances the model's ability to extract modality-specific information, leading to more accurate representations.
We also find that separating the self-attention layer and sharing the FFN layer yields the poorest results. We speculate that this is due to the self-attention layer playing a more significant role in aligning modalities compared to the FFN layer. Therefore, separating the self-attention layer lead to inferior performance.
Figure~\ref{fig:ablation_shared} demonstrates the convergence performance of different architectures. 
Among all the architectures, the model with shared self-attention layers and separated FFNs exhibits the fastest convergence speed.

\paragraph{Effects of Intra-modal Denoising Contrastive Learning.}
We examine the effects of intra-modal denoising contrastive learning (DCL).
As shown in Table~\ref{tb:loss_ablation}, applying DCL to language data (DCL-L) can enhance the model's performance in text retrieval tasks.
Furthermore, applying DCL to vision data (DCL-V) can improve the model's cross-modal retrieval ability, as well as fine-tuning performance in image classification. 
By applying DCL to vision-language data (DCL-VL), \onepeace achieves the best results in terms of all the evaluation metrics. 
These results demonstrate that intra-modal denoising contrastive learning can complement cross-modal contrastive learning.
It not only enables \onepeace to achieve excellent downstream fine-tuning performance but also enhances the model's capability for zero-shot cross-modal retrieval.

\paragraph{Ablation on Different Denoising Losses.}
We conduct a systematic comparison of different denoising losses, including the smooth L1 loss used in~\cite{data2vec,maskclip}, the L2 loss used in~\cite{data2vec2,dbot}, the cosine loss used in~\cite{mvp,eva,caev2}, and the denoising contrastive loss used in this paper. 
As shown in Table~\ref{tb:masked_loss_ablation}, different types of denoising loss can improve the performance of the model in both cross-modal retrieval and image classification tasks. 
Among all the denoising losses, the denoising contrastive loss has the greatest improvement in terms of all the metrics compared to other losses. For example, it increased by $+1.64$ on COCO text retrieval R@1, increased by $+1.47$ on COCO image retrieval R@1, and increased by $+0.6$ on image classification. 
This indicates that denoising contrastive loss is more compatible with cross-modal contrastive loss than other denoising losses.

\subsection{Emergent Zero-shot Retrieval}
In our pretraining, we exclusively align other modalities with text which plays as an intermediary role. We assume that our model is able to align those modalities that are not paired in the training data. For example, \onepeace~should be able to align image and audio. Thus, we conduct experiments on the retrieval of those modalities to assess the emergent zero-shot capabilities~\cite{imagebind}. 

\begin{figure*}[t]
\vskip 0.2in
    \centering
    \includegraphics[width=1.0\linewidth]{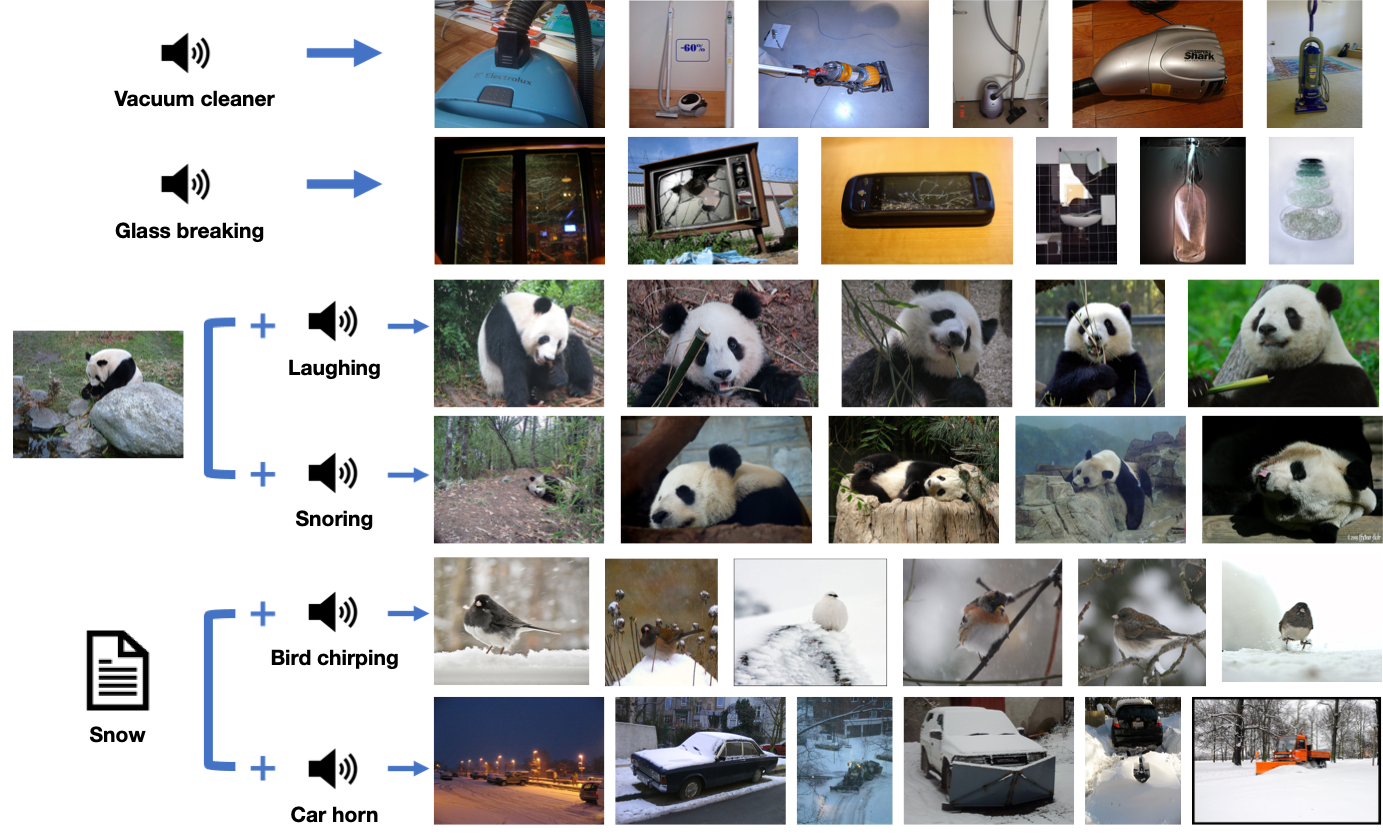}
    \caption{Examples of emergent zero-shot retrieval. \onepeace~is capable of aligning modalities and modality combinations, where there are no paired data of the modalities in the pretraining dataset. The images are retrieved from ImageNet-1K and MSCOCO.}
    \label{fig:case1}
\end{figure*}

To be more specific, we evaluate the audio-to-image, audio+image-to-image, and audio+text-to-image retrieval abilities and demonstrate case studies in Figure~\ref{fig:case1}. 
The first two cases demonstrate the emergent capability of uni-modal retrieval, while the other cases show that of the retrieval of image based on multimodal inputs. 
Specifically, we find that \onepeace~is able to retrieve images that contain elements concerning inputs of different modalities, e.g., the model uses the text ``snow'' and the sound of bird chirping to retrieve the images of birds in the snow. These examples demonstrate that \onepeace~has strong potential in emergent zero-shot capabilities. This indicates that for a universal representation model, there is no need to learn all pairing relationships between modalities, but instead it is sufficient for modalities to be aligned to an intermediary one. 
We provide more quality examples in Appendix \ref{app:emergent_zero_shot}.

%% file: table/v_result.tex

\begin{table*}[t]
\centering
\normaltablestyle{6pt}{1.2}
\begin{tabular}{y{85}|cc|cc}
Method                                 & Enc. \#Params & Crop Size & mIoU$^{ss}$ & mIoU$^{ms}$ \\
\shline
RevCol-H~\cite{cai2022reversible}      & 2.16B    & 640$^2$   & 60.4        & 61.0 \\
FD-SwinV2-G~\cite{wei2022contrastive}  & 3.00B    & 896$^2$   & -           & 61.4 \\
ViT-Adapter~\cite{chen2022vitadapter}  & 571M     & 896$^2$   & 61.2        & 61.5 \\
EVA~\cite{eva}                         & 1.01B    & 896$^2$   & 61.5        & 62.3 \\
BEiT-3~\cite{beit3}                    & 1.01B    & 896$^2$   & 62.0        & 62.8 \\
InternImage~\cite{wang2022internimage} & 1.08B    & 896$^2$   & \textbf{62.5}        & 62.9 \\
\hline
\modelname                             & 1.52B    & 896$^2$   & 62.0  & \textbf{63.0}
\end{tabular}
\caption{\textbf{System-level comparisons of semantic segmentation with leading results on ADE20k.} mIoU$^{ss}$ means single-scale inference result while mIoU$^{ms}$ means multi-scale.}
\label{tb:ade20k}
\end{table*}

\begin{table*}[t]
\centering
\normaltablestyle{6pt}{1.2}
\begin{tabular}{y{80}|ccc|cc}
Method                                & Detector & \#Params & Image Size & AP$^{box}$ & AP$^{mask}$ \\
\shline
ViT-Adapter~\cite{chen2022vitadapter} & HTC++    & 401M     & [400-1400, 1600] & 58.8    & 51.1 \\
ViTDet~\cite{vitdet}               & Cascade    & 692M     & 1280$^2$ & 60.4    & 52.0 \\
RevCol-H~\cite{cai2022reversible}  & HTC++    & 2.41B    & [400-1400, 1600] & \textbf{61.1}    & \textbf{53.0} \\
\hline
\modelname                         & Cascade  & 1.59B    & 1280$^2$    & 60.4 & 52.9
\end{tabular}
\caption{\textbf{System-level comparisons of object detection and instance segmentation on MSCOCO.} The reported results are obtained by directly fine-tuning the models on MSCOCO, without intermediate fine-tuning on Objects365.}
\label{tb:coco}
\vspace{+1em}
\end{table*}

\begin{table*}[t]
\centering
\normaltablestyle{6pt}{1.2}
\begin{tabular}{*l| ^c ^c ^c ^c}
Method & Backbone & Input Size & Top-1 & Top-5
\\
\shline
VATT~\cite{vatt} & ViT-L    & 32 $\times$ 320$^2$ & 82.1 & 95.5
\\
ViViT~\cite{vivit} & ViT-H    & 32 $\times$ 224$^2$ & 84.9 & 95.8
\\
Florance~\cite{florence} & Co-Swin-H & N/A $\times$ 384$^2$ & 86.5 & 97.3
\\
SwinV2~\cite{swinv2} & Swin-G & N/A $\times$ 384$^2$ & 86.8 & -
\\
MAE-ST~\cite{maest} & ViT-H & 16 $\times$ 224$^2$ & 86.8 & 97.2
\\
VideoMAE~\cite{videomae} & ViT-H & 32 $\times$ 320$^2$ & 87.4 & 97.6
\\
VideoMAE V2~\cite{videomaev2} & ViT-H & 32 $\times$ 320$^2$ & 87.4 & 97.6
\\
MaskFeat~\cite{maskfeat} & MViTv2-L & 40 $\times$ 352$^2$ & 87.0 & 97.4
\\
CoCa (frozen)~\cite{coca} & ViT-g & 16 $\times$ 576$^2$ & 88.0 & -
\\
ViT-22B (frozen)~\cite{vit22b} & ViT-22B & 128 $\times$ 224$^2$ & 88.0 & -
\\
\hline
\modelname (frozen) & ViT-g & 16 $\times$ 256$^2$ & 88.0 & \textbf{97.8}
\\
\modelname (frozen) & ViT-g & 32 $\times$ 256$^2$ & \textbf{88.1} & \textbf{97.8}
\\
\hline
\multicolumn{5}{l}{\rowstyle{\color{dt}}\textit{methods using intermediate fine-tuning:}}
\\
\hline
\rowstyle{\color{dt}}EVA~\cite{eva} & \color{dt}ViT-g & 16 $\times$ 224$^2$ & 89.7 & -
\\
\rowstyle{\color{dt}}VideoMAE V2~\cite{videomaev2} & ViT-g & 64 $\times$ 266$^2$ & 90.0 & 98
\\
\end{tabular}
\caption{\textbf{System-level comparisons of video action recognition with leading results on Kinetics-400.} Frozen means do not update pre-trained model parameters. EVA~\cite{eva} and VideoMAE V2~\cite{videomaev2} are intermediate fine-tuned on the merged Kinetics dataset (K400, K600, K700).}
\label{tb:k400}
\end{table*}

%% file: table/audio_result.tex
\begin{table*}[t]
\centering
\normaltablestyle{4pt}{1.2}
\begin{tabular}{l|cccccc|cccccc}
  \multirow{3}*{Method}
  &\multicolumn{6}{c|}{AudioCaps}
  &\multicolumn{6}{c}{Clotho}
  \\
  &\multicolumn{3}{c}{Text $\rightarrow$ Audio}
  &\multicolumn{3}{c|}{Audio $\rightarrow$ Text}
  &\multicolumn{3}{c}{Text $\rightarrow$ Audio}
  &\multicolumn{3}{c}{Audio $\rightarrow$ Text}
  \\
  & \footnotesize R@1 & \footnotesize R@5 & \footnotesize R@10 & \footnotesize R@1 & \footnotesize R@5 & \footnotesize R@10 & \footnotesize R@1 & \footnotesize R@5 & \footnotesize R@10 & \footnotesize R@1 & \footnotesize R@5 & \footnotesize R@10
  \\
  \shline
  MMT \cite{MMT}
  & 36.1 & 72.0 & 84.5
  & 39.6 & 76.8 & 86.7
  & 6.7 & 21.6 & 33.2
  & 7.0 & 22.7 & 34.6
  \\
  ML-ACT \cite{ML-ACT}
  & 33.9 & 69.7 & 82.6
  & 39.4 & 72.0 & 83.9
  & 14.4 & 36.6 & 49.9
  & 16.2 & 37.6 & 50.2
  \\
  CLAP-HTSAT \cite{wavtext5k}
  & 34.6 & 70.2 & 82.0
  & 41.9 & 73.1 & 84.6
  & 16.7 & 41.1 & 54.1
  & 20.0 & 44.9 & 58.7
  \\
  TAP \cite{TAP}
  & 36.1 & 72.0 & 85.2
  & 41.3 & 75.5 & 86.1
  & 16.2 & 39.2 & 50.8
  & 17.6 & 39.6 & 51.4
  \\
  LAION-CLAP \cite{laion_clap}
  & 35.1 & 71.5 & 83.6
  & 45.8 & 80.9 & 91.6
  & 18.2 & 42.5 & 54.4
  & 25.7 & 51.5 & 63.4
  \\
  \hline
  \modelname
  & \textbf{42.5} & \textbf{77.5} & \textbf{88.4}
  & \textbf{51.0} & \textbf{81.9} & \textbf{92.0}
  & \textbf{22.4} & \textbf{49.0} & \textbf{62.7}
  & \textbf{27.1} & \textbf{52.3} & \textbf{65.4}
  \\
\end{tabular}
\caption{\textbf{Experimental results on audio-text retrieval.} \onepeace significantly outperforms baselines by a large margin.}
\label{tb:audio_text_retrieval}
\end{table*}

\begin{table*}[t]
\centering
\normaltablestyle{8pt}{1.3}
\begin{tabular}{l|cccc}
  \multirow{2}*{Method}
  &\multicolumn{1}{c}{ESC-50}
  &\multicolumn{1}{c}{FSD50K}
  &\multicolumn{1}{c}{VGGSound (Audio Only)}
  &\multicolumn{1}{c}{AQA}
  \\
  &\multicolumn{1}{c}{ZS}
  &\multicolumn{1}{c}{FT}
  &\multicolumn{1}{c}{FT}
  &\multicolumn{1}{c}{FT}
  \\
  \shline
  Previous SOTA
  & 91.0 \cite{laion_clap} & 65.6 \cite{PaSST} & 59.5 \cite{cav_mae} & 83.5 \cite{avqa}
  \\
  \hline
  Wav2CLIP \cite{align}
  & 41.4 & 43.1 & 46.6 & -
  \\
  AudioCLIP \cite{filip}
  & 69.4 & - & - & -
  \\
  CLAP \cite{clap}
  & 82.6 & 58.6 & - & -
  \\
  LAION-CLAP \cite{laion_clap}
  & 91.0 & 46.2 & 55.1* & -
  \\
  \hline
  \modelname
  & \textbf{91.8} & \textbf{69.7} & \textbf{59.6} & \textbf{86.2}
  \\
\end{tabular}
\caption{\textbf{Experimental results on audio classification and audio question answering (AQA).} "ZS" is short for zero-shot results, "FT" is short for fine-tuning results. For the VGGSound dataset, we only use the audio data and discard the video data. *We use the official code of LAION-CLAP to reproduce the result on VGGSound.}
\label{tb:audio_result}
\end{table*}


%% file: table/image_text_result.tex
\begin{table*}[t]
\centering
\normaltablestyle{4pt}{1.2}
\begin{tabular}{lcccccc|cccccc}
  \multirow{3}*{Method}
  &\multicolumn{6}{c|}{COCO (5K test set)}
  &\multicolumn{6}{c}{Flickr30K (1K test set)}
  \\
  &\multicolumn{3}{c}{Image $\rightarrow$ Text}
  &\multicolumn{3}{c|}{Text $\rightarrow$ Image}
  &\multicolumn{3}{c}{Image $\rightarrow$ Text}
  &\multicolumn{3}{c}{Text $\rightarrow$ Image}
  \\
  & \footnotesize R@1 & \footnotesize R@5 & \footnotesize R@10 & \footnotesize R@1 & \footnotesize R@5 & \footnotesize R@10 & \footnotesize R@1 & \footnotesize R@5 & \footnotesize R@10 & \footnotesize R@1 & \footnotesize R@5 & \footnotesize R@10
  \\
  \shline
  \multicolumn{13}{l}{\textit{Zero-shot Setting}}
  \\
  CLIP~\cite{clip}
  & 58.4 & 81.5 & 88.1
  & 37.8 & 62.4 & 72.2
  & 88.0 & 98.7 & 99.4
  & 68.7 & 90.6 & 95.2
  \\
  ALIGN~\cite{align}
  & 58.6 & 83.0 & 89.7
  & 45.6 & 69.8 & 78.6
  & 88.6 & 98.7 & 99.7
  & 75.7 & 93.8 & 96.8
  \\
  FILIP~\cite{filip}
  & 61.3 & 84.3 & 90.4
  & 45.9 & 70.6 & 79.3
  & 89.8 & 99.2 & 99.8
  & 75.0 & 93.4 & 96.3
  \\
  Florence~\cite{florence}
  & 64.7 & 85.9 & -
  & 47.2 & 71.4 & -
  & 90.9 & 99.1 & -
  & 76.7 & 93.6 & -
  \\
  CoCa~\cite{coca}
  & \textbf{66.3} & \textbf{86.2} & 91.8
  & \textbf{51.2} & \textbf{74.2} & \textbf{82.0}
  & \textbf{92.5} & \textbf{99.5} & \textbf{99.9}
  & \textbf{80.4} & \textbf{95.7} & \textbf{97.7}
  \\
  \modelname
  & 64.7 & 86.0 & \textbf{91.9}
  & 48.0 & 71.5 & 79.6
  & 90.9 & 98.8 & 99.8
  & 77.2 & 93.5 & 96.2
  \\
  \shline
  \multicolumn{13}{l}{\textit{Fine-tuning Setting}}
  \\
  ALIGN~\cite{align}
  & 77.0 & 93.5 & 96.9
  & 59.9 & 83.3 & 89.8
  & 95.3 & 99.8 & 100.0
  & 84.9 & 97.4 & 98.6
  \\
  FILIP~\cite{filip}
  & 78.9 & 94.4 & 97.4
  & 61.2 & 84.3 & 90.6
  & 96.6 & 100.0 & 100.0
  & 87.1 & 97.7 & 99.1
  \\
  Florence~\cite{florence}
  & 81.8 & 95.2 & -
  & 63.2 & 85.7 & -
  & 97.2 & 99.9 & -
  & 87.9 & 98.1 & -
  \\
  OmniVL~\cite{omnivl}
  & 82.1 & 95.9 & 98.1
  & 64.8 & 86.1 & 91.6
  & 97.3 & 99.9 & \textbf{100.0}
  & 87.9 & 97.8 & 99.1
  \\
  BEiT-3~\cite{beit3}
  & 82.7 & 96.0 & 98.2
  & 65.1 & \textbf{86.6} & \textbf{92.3}
  & 97.5 & 99.9 & \textbf{100.0}
  & 89.1 & \textbf{98.6} & \textbf{99.3}
  \\
  \modelname
  & \textbf{84.1} & \textbf{96.3} & \textbf{98.3}
  & \textbf{65.4} & 86.3 & 91.9
  &\textbf{97.6} & \textbf{100.0} & \textbf{100.0}
  &\textbf{89.6} & 98.0 & 99.1
  \\
\end{tabular}
\caption{\textbf{Experimental results on image-text retrieval.} We compare with baselines under both zero-shot and fine-tuning settings. For a fair comparison, the reported results of BEiT-3 are obtained by directly fine-tuning on downstream benchmarks without intermediate fine-tuning on pretraining data.}
\label{tb:vl-zeroshot}
\end{table*}

%% file: table/vl_result.tex
\begin{table*}[t]
\centering
\normaltablestyle{8pt}{1.2}
\begin{tabular}{l|ccc|ccc|cc}
  \multirow{2}*{Method}
  & \multicolumn{3}{c|}{RefCOCO} & \multicolumn{3}{c|}{RefCOCO+} & \multicolumn{2}{c}{RefCOCOg}
  \\
  & val & testA & testB & val & testA & testB & val-u & test-u
  \\
  \shline
  VL-T5 \cite{vlt5}
  &- & - & -
  &- & - & -
  &- & 71.3
  \\
  UNITER \cite{uniter}
  &81.41 & 87.04 & 74.17
  &75.90 & 81.45 & 66.70
  &74.86 & 75.77
  \\
  VILLA \cite{villa}
  &82.39 & 87.48 & 74.84
  &76.17 & 81.54 & 66.84
  &76.18 & 76.71
  \\
  MDETR \cite{Kamath2021MDETRM}
  &86.75 & 89.58 & 81.41
  &79.52 & 84.09 & 70.62
  &81.64 & 80.89
  \\
  UNICORN \cite{unicorn}
  &88.29 & 90.42 & 83.06
  &80.30 & 85.05 & 71.88
  &83.44 & 83.93
  \\
  X-VLM \cite{x-vlm}
  & - & - & -
  & 84.51 & 89.00 & 76.91
  & - & -
  \\
  Grounding-DINO \cite{grounding_dino}
  & 90.56 & 93.19 & 88.24
  & 82.75 & 88.95 & 75.92
  & 86.13 & 87.02
  \\
  FIBER \cite{fiber}
  & 90.68 & 92.59 & 87.26
  & 85.74 & 90.13 & 79.38
  & 87.11 & 87.32
  \\
  OFA \cite{ofa}
  & 92.04 & 94.03 & 88.44
  & 87.86 & 91.70 & 80.71
  & 88.07 & 88.78
  \\
  \hline
  \modelname
  & \textbf{92.58} & \textbf{94.18} & \textbf{89.26}
  & \textbf{88.77} & \textbf{92.21} & \textbf{83.23}
  & \textbf{89.22} & \textbf{89.27}
  \\
\end{tabular}
\caption{Experimental results on $3$ visual grounding datasets: RefCOCO, RefCOCO+, RefCOCOg. \onepeace achieves state-of-the-are results without using additional visual grounding datasets (e.g., Visual genome).}
\label{tb:refcoco}
\end{table*}

%% file: table/vision_language_understanding.tex
\begin{table*}[t]
\centering
\normaltablestyle{8pt}{1.3}
\begin{tabular}{l|cc|cc}
  \multirow{2}*{Method}
  & \multicolumn{2}{c|}{VQA} & \multicolumn{2}{c}{NLVR-2}
  \\
  & test-dev & test-std & dev & test-P
  \\
  \shline
  ALBEF \cite{albef}
  & 75.8 & 76.0 & 82.55 & 83.14
  \\
  BLIP \cite{blip} 
  & 78.25 & 78.32 & 82.15 & 82.24
  \\
  X-VLM \cite{x-vlm}
  & 78.22 & 78.37 & 84.41 & 84.76
  \\
  SimVLM \cite{simvlm}
  & 80.0 & 80.3 & 84.5 & 85.2
  \\
  OFA \cite{ofa}
  & 82.0 & 82.0 & - & -
  \\
  Flamingo \cite{flamingo}
  & 82.0 & 82.1 & - & -
  \\
  CoCa \cite{coca}
  & 82.3 & 82.3 &86.1 & 87.0
  \\
  BLIP-2 \cite{blip2}
  & 82.2 & 82.3 & - & -
  \\
  BEiT-3 \cite{beit3}
  & \textbf{84.2} & \textbf{84.0} & \textbf{91.5} & \textbf{92.6}
  \\
  \hline
  \modelname
  & 82.6 & 82.5 & 87.8 & 88.3
  \\
\end{tabular}
\caption{\textbf{Results on vision-language understanding tasks.} Without initialized with language pretrained models or pretraining on pure text data, \onepeace outperforms the strong baselines Flamingo and CoCa.}
\label{tb:vqa-nlvr-ve}
\end{table*}

%% file: table/ablation.tex
\begin{table*}[t]
\centering
\normaltablestyle{6pt}{1.2}
\begin{tabular}{l|ccc|ccc|c}
  \multirow{3}*{Structure}
  & \multicolumn{6}{c|}{COCO zero-shot (5k test set)} & \multirow{2}*{IN-1K}
  \\
  & \multicolumn{3}{c|}{Image $\rightarrow$ Text} & \multicolumn{3}{c|}{Text $\rightarrow$ Image} &
  \\
  & \footnotesize R@1 & \footnotesize R@5 & \footnotesize R@10 & \footnotesize R@1 & \footnotesize R@5 & \footnotesize R@10 & ZS
  \\
  \shline
  Share ATTN \& FFN & 33.96 & 60.92 & 72.24 & 22.51 & 46.13 & 57.63 & 42.21
  \\ 
  No Share & 29.08 & 54.30 & 65.56 & 18.98 & 40.34 & 51.44 & 37.73
  \\
  Share FFN & 27.36 & 52.66 & 64.36 & 18.08 & 38.68 & 49.84 & 33.98
  \\
  Share ATTN & \textbf{35.94} & \textbf{62.52} & \textbf{72.78} & \textbf{23.87} & \textbf{47.80} & \textbf{59.38} & \textbf{43.24}
  \\
\end{tabular}
\caption{\textbf{Ablation experiments on model structures.} "Share ATTN \& FFN" means share both self-attention layers and FFN layers. "No Share" means separate both self-attention layers and FFN layers. "Share FFN" means separate self-attention layers and share FFN layers. "Share ATTN" means share self-attention layers and separate FFN layers, which is the default setting of \onepeace. "ZS" is short for zero-shot accuracy.}
\label{tb:structure_ablation}
\vspace{+1em}
\end{table*}

\begin{figure*}[t]\centering
\newcommand{\hei}{0.33\linewidth}
\subfloat[\textbf{Training loss}\label{pic:ablation_loss}]
{\includegraphics[width=\hei,height=0.27\linewidth]{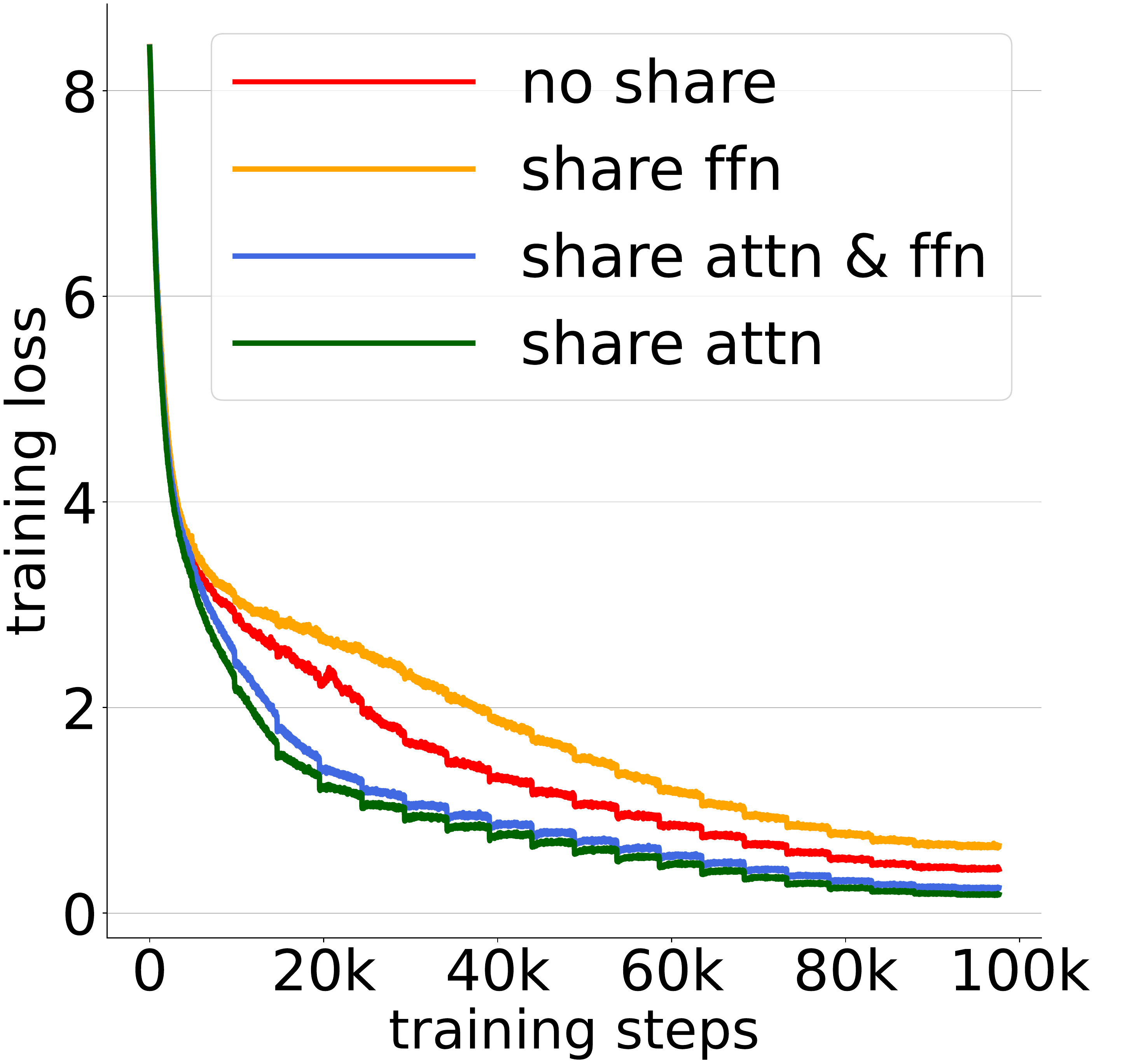}}
\subfloat[\textbf{Training accuracy}\label{pic:ablation_batch_acc}]
{\includegraphics[width=\hei,height=0.27\linewidth]{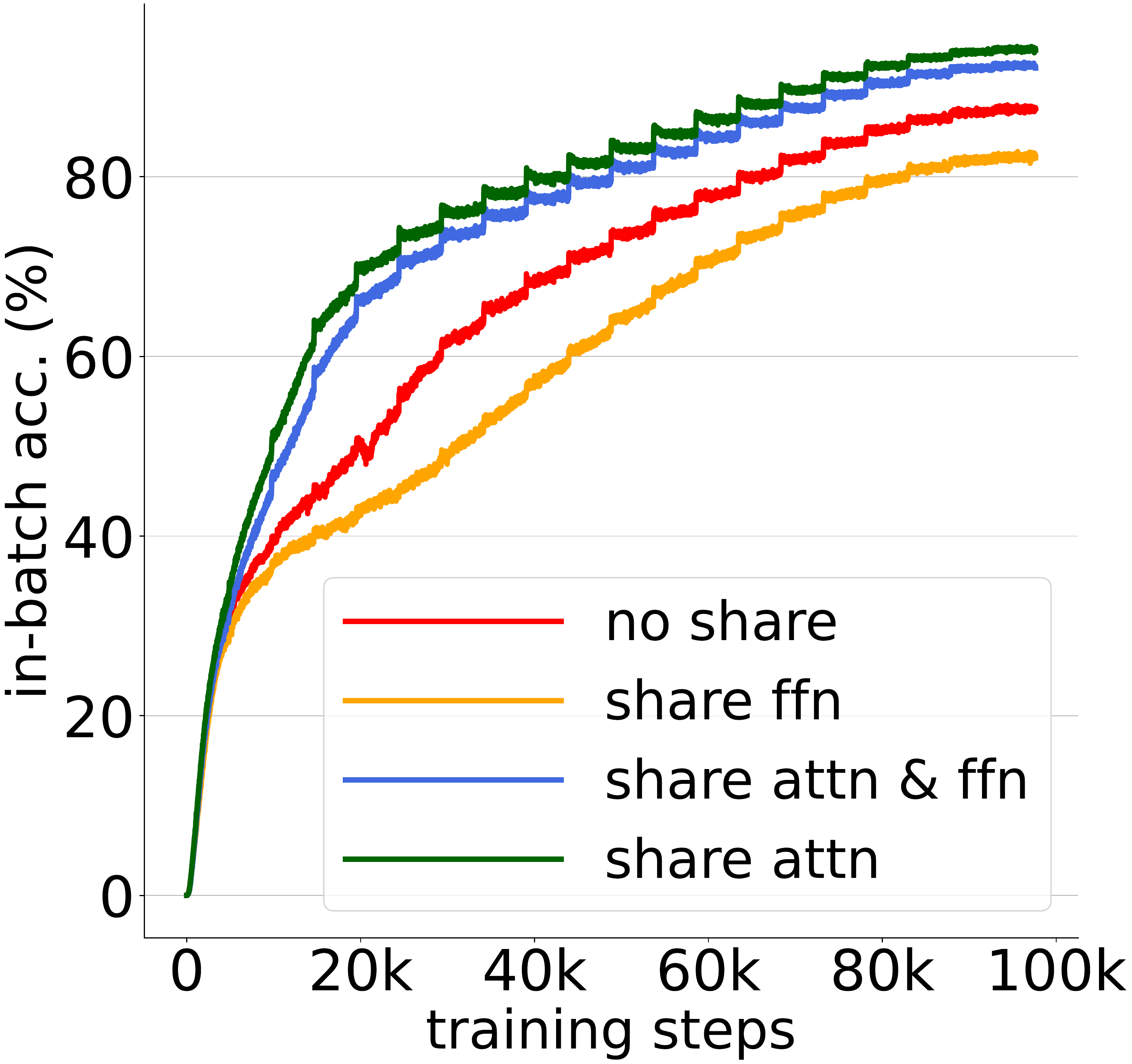}}
\subfloat[\textbf{Zero-shot accuracy} \label{pic:ablation_epoch_acc}]
{\includegraphics[width=\hei,height=0.27\linewidth]{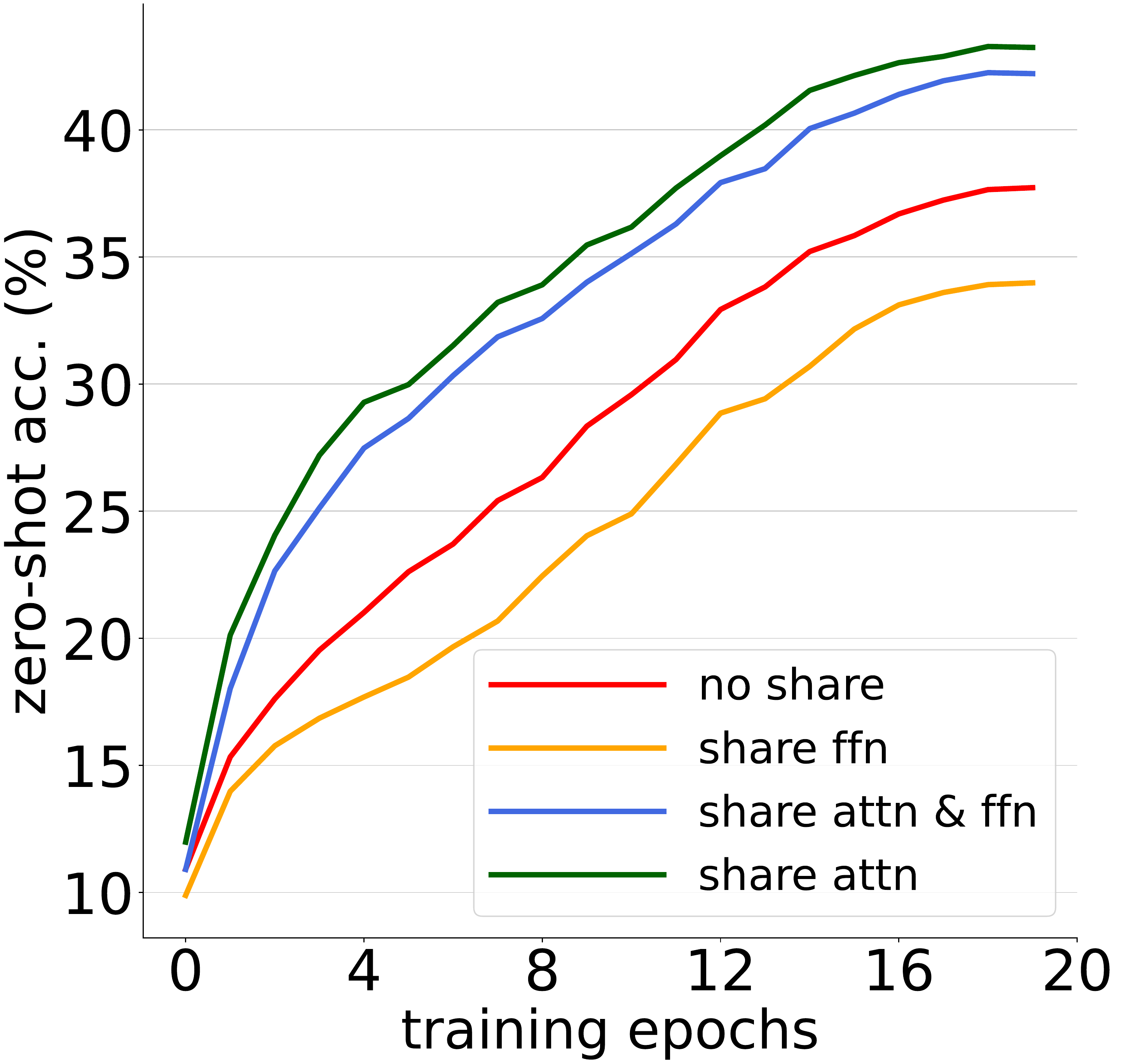}}
\caption{\textbf{Training curves of different structures.} The model with shared self-attention layers and separated FFNs ("share attn") outperforms other structures, exhibiting the fastest convergence speed. (The curve appears as a staircase shape because we use exponential moving average to calculate relevant indicators in each epoch.)}
\label{fig:ablation_shared}
\end{figure*}

\begin{table*}[t]
\centering
\normaltablestyle{6pt}{1.2}
\begin{tabular}{cccc|ccc|ccc|cc}
  & & & &\multicolumn{6}{c|}{COCO zero-shot (5k test set)} & \multicolumn{2}{c}{\multirow{2}*{IN-1K}}
  \\
  & & & &\multicolumn{3}{c|}{Image $\rightarrow$ Text} &\multicolumn{3}{c|}{Text $\rightarrow$ Image}
  \\
  \footnotesize CL & \footnotesize DCL-L & \footnotesize DCL-V & \footnotesize DCL-VL & \footnotesize R@1 & \footnotesize R@5 & \footnotesize R@10 & \footnotesize R@1 & \footnotesize R@5 & \footnotesize R@10 & \footnotesize ZS & \footnotesize FT
  \\
\shline
  \checkmark & & & & 35.94 & 62.52 & 72.78 & 23.87 & 47.80 & 59.38 & 43.24 & 82.20
  \\ 
  \checkmark & \checkmark & & & 37.02 & 63.78 & 73.96 & 23.63 & 47.87 & 59.22 & 43.69 & 81.99
  \\
  \checkmark & & \checkmark & & 38.88 & 65.34 & 75.76 & 26.05 & 49.78 & 61.26 & 45.94 & 83.32
  \\
  \checkmark & \checkmark & \checkmark & & 39.00 & 65.64 & 76.30 & 25.85 & 50.06 & 61.80 & 45.54 & 83.33
  \\
  \checkmark & \checkmark & \checkmark & \checkmark & \textbf{39.94} & \textbf{65.94} & \textbf{76.72} & \textbf{26.94}  & \textbf{51.38} & \textbf{62.81} & \textbf{46.41} & \textbf{83.75}
  \\
\end{tabular}
\caption{\textbf{Ablation studies of intra-modal denoising contrastive learning.} "CL" is cross-modal contrastive learning. "DCL-L", "DCL-V", and "DCL-VL" means applying intra-modal denoising contrastive learning to language, vision, and vision-language data, respectively.  "ZS" is short for zero-shot accuracy, "FT" is short for fine-tuning accuracy.}
\label{tb:loss_ablation}
\vspace{+1em}
\end{table*}

\begin{table*}[t]
\centering
\normaltablestyle{6pt}{1.2}
\begin{tabular}{*l| ^c ^c ^c| ^c ^c ^c| ^c ^c ^c}
  & \multicolumn{6}{c|}{COCO zero-shot (5k test set)} & \multicolumn{2}{c}{\multirow{2}*{IN-1K}}
  \\
  & \multicolumn{3}{c|}{Image $\rightarrow$ Text} & \multicolumn{3}{c|}{Text $\rightarrow$ Image} &
  \\
  Denoising Loss & \footnotesize R@1 & \footnotesize R@5 
  & \footnotesize R@10
  & \footnotesize R@1 & \footnotesize R@5 & \footnotesize R@10
  & ZS & FT
  \\
  \shline
  None & 35.94 & 62.52 & 72.78 & 23.87 & 47.80 & 59.38 & 43.24 & 82.20
  \\
  Smooth L1 Loss & 36.72 & 63.86 & 74.30 & 24.37 & 48.17 & 59.86 & 44.36 & 82.50
  \\
  L2 Loss & 37.46 & 64.80 & 74.46 & 25.81 &49.64 & 61.80 & 45.78 & 83.15
  \\
  Cosine Loss & 38.28 & 65.80 & 75.18 & 25.47 & 50.02 & 61.59 & 45.15 & 83.07
  \\ 
  Denoising Contrastive Loss & \textbf{39.94} & \textbf{65.94} &
  \textbf{76.72} &
  \textbf{26.94} & \textbf{51.38} & 
  \textbf{62.81} &
  \textbf{46.41} & \textbf{83.75}
  \\
\end{tabular}
\caption{\textbf{Ablation studies of different denoising losses.} Among all the denoising losses, denoising contrastive loss shows the greatest performance improvement in both cross-modal retrieval and image classification tasks.}
\label{tb:masked_loss_ablation}
\vspace{+1em}
\end{table*}


%% file: content/5_conclusion.tex
\section{Conclusion, Limitation and Future Work}
In this work, we explore a scalable way for building a general representation model across different modalities.
Based on the flexible architecture and modality-agnostic pretraining tasks, we release \onepeace, a general representation model that can seamlessly align and integrate representations across vision, audio, and language modalities.
We conduct a series of experiments across 3 modalities, 11 tasks, and 16 datasets.
The experimental results demonstrate that \onepeace achieves leading results in a wide range of tasks, including image classification, semantic segmentation, audio-text retrieval, audio classification, audio question answering, image-text retrieval, and visual grounding.
Furthermore, we show that \onepeace possesses a strong emergent zero-shot retrieval capability, enabling it to align modalities that are not paired in the training data.

\paragraph{Limitation.}
Although \onepeace achieves leading results in a wide range of tasks, it falls short of achieving state-of-the-art results in zero-shot image-text retrieval and vision-language understanding tasks.
There are two possible reasons for this:
1). \onepeace didn't see enough image-text pairs during pretraining. We only trained on 6.4 billion image-text pairs, while previous works~\cite{clip,align,coca} typically train on 12.8 billion image-text pairs or more.
2). \onepeace didn't use language pretrained models for initialization or introduce any pure text data. Both the vision and language modules of \onepeace are completely randomly initialized, while previous works~\cite{ofa,blip2} show that introducing pure text data or initialized with the language pretrained models can greatly enhance the model's performance. 
In fact, as a highly extensible model, \onepeace can combine with language pretrained models to achieve better results.

\paragraph{Future Work.}
In the future, we will test \onepeace on more downstream tasks, such as vision-audio-language tasks and extend to more modalities for pretraining like video, 3D point cloud, etc.
Also, we pursue an active interaction with large language models (LLMs) to continue influencing broader areas. This includes:

\begin{itemize}
    \item With the help of LLMs, building a more powerful general representation model.
    \item By combining LLMs, creating a more general multimodal language model.
\end{itemize}

%% file: content/6_appendix.tex
\section{Pretraining Details}

\subsection{Pretraining Datasets}
\label{app:audio_text_data_details}
For image-text pairs, we use LAION-2B~\cite{laion5b}, a dataset obtained by web crawling that may contain some noisy pairs. 
To improve the data quality, we apply several pre-processing steps, including removing images with an aspect ratio greater than 3.5, removing images with the shortest side less than 128, and removing images with a CLIP score less than 0.3. We also remove texts containing non-English or emoji characters, as well as texts with lengths less than 3 or greater than 512. 
After these steps, we retain about 1.5 billion image-text pairs.

For audio-text pairs, we mainly use the environmental sound datasets processed by \cite{laion_clap}. Specifically, for some datasets that only contain tags, \cite{laion_clap} uses a pretrained language model T5~\cite{T5} to rewrite these tags into captions.
We also perform simple cleaning on the data, which involves removing samples with text lengths less than 3 or greater than 512, as well as texts containing non-English or emoji characters. 
Ultimately, we obtain about 2.4 million audio-text pairs, with a total duration of around 8,000 hours.
Table~\ref{tb:audio_dataset} presents the environmental sound datasets utilized by \onepeace.

\input{table/audio_dataset}

\subsection{Pretraining Settings}
\label{app:pretraining_hyperparameters}




As mentioned in Sec~\ref{sec:two_stage_pretraining}, the pretraining of \onepeace is divided into two stages: vision-language pretraining and audio-language pretraining.

For vision-language pretraining, we pretrain \onepeace for 200K steps with a batch size of 32768. We use the AdamW~\cite{adamw} optimizer with $(\beta_1,\beta_2)=(0.9, 0.98)$ and $\epsilon=1e\text{-}8$. The peak learning rate is set to $5e-4$, with a linear warmup of $3000$ steps and a cosine decay scheduler. The image resolution is set to $256\times256$. The maximum text sequence length is set to 70. For regulation, we use weight decay with $0.05$ and disable dropout. We employ drop path~\cite{drop-path} with a $0.4$ rate.

For audio-language pretraining, we keep the model parameters related to vision and language (e.g., self-attention layers) frozen and only update the parameters that pertain to audio, such as A-Adapter and A-FFN.
In this stage, we pretrain \onepeace for $10$ epochs with a batch size of $3072$. The peak learning rate is set to $2e-4$, with a linear warmup of $1$ epoch and cosine decay scheduler. The maximum audio duration is set to $15$s. For audio with a duration of less than $1$s, we first repeat the input and then truncate it to $1$s.
Other hyper-parameters remain the same as vision-language pretraining.



\section{Details of Downstream Tasks}
\label{app:downstream_tasks}

\subsection{Vision Tasks}
\label{app:vision_details}
Here we describe the implementation details of different vision tasks, including image classification~\cite{inet1k}, semantic segmentation~\cite{Zhou2016SemanticUO}, object detection~\cite{mscoco}, and video action recognition~\cite{k400}.
All detailed hyperparameters are listed in Table~\ref{tb:v_task_config}.

\paragraph{Image Classification}
We provide the fine-tuning results on ImageNet-1k~\cite{inet1k}. Following recent studies in self-supervised learning for computer vision, we use global pooling of all image tokens excluding the class token, and append a LayerNorm with a linear layer for classification. To further unleash the potential of \onepeace, we perform intermediate fine-tuning on ImageNet-21k~\cite{imagenet}. We set the label smoothing as 0.3 and do not use random erasing, mixup, and cutmix data augmentations. For fine-tuning on ImageNet-1k, we use exponential moving average (EMA) for model parameters and set the EMA decay rate as 0.9998. For intermediate fine-tuning on ImageNet-21k, we do not use EMA. We also use Zero Redundancy Optimizer~\cite{zero} and set the stage as 1.

\paragraph{Semantic Segmentation}
We provide the fine-tuning results on ADE20k~\cite{Zhou2016SemanticUO}. We use Mask2Former~\cite{mask2former} as the segmentation head. We first intermediate fine-tune segmentation head on coco-stuff~\cite{cocostuff} dataset for 80k steps. The learning rate is set as 2e-5 and the rest hyperparameters are the same as ADE20K shown in Table~\ref{tb:v_task_config}. Then we fine-tune the model on ADE20K. Both experiments use the cosine learning rate decay scheduler.

\paragraph{Object Detection}
We provide the fine-tuning results on COCO~\cite{mscoco} with ViTDet~\cite{vitdet}. We use large-scale jitter~\cite{lsj} data augmentation and fine-tune for 50 epochs. We use the linear learning rate decay scheduler and decay the learning rate at 44 and 48 epochs respectively.

\paragraph{Video Action Recognition}
To perform video action recognition, following AIM~\cite{aim}, we freeze the parameters of the pre-trained model and add spatial and temporal MLP adapters in each transformer layer. We conduct experiments on Kinetics 400~\cite{k400} dataset. Due to the invalid video links, there are many different versions of the K400 dataset and we use the version released on AcademicTorrents. We use the cosine learning decay scheduler and set the backbone learning rate multiplier of 0.1.

\begin{table*}[t]
\normaltablestyle{8pt}{1.2}
\begin{tabular}{y{96}|ccccc}
Config & ImageNet-21k & ImageNet-1k & ADE20K & COCO & Kinetics 400\\
\shline
Optimizer & \multicolumn{5}{c}{AdamW} \\
Optimizer momentum & \multicolumn{5}{c}{$\beta_1, \beta_2{=}0.9, 0.999$} \\
Numerical precision & \multicolumn{5}{c}{$\mathtt{fp16}$} \\
Peak learning rate & 1e-4 & 5e-5 & 1.5e-5 & 1e-4 & 3e-4 \\
Layer-wise lr decay & 0.85 & 0.9 & 0.95 & 0.9 & - \\
Weight decay & 0.05 & 0.05 & 0.05 & 0.1 & 0.05 \\
Batch size & 5120 & 1024 & 16 & 64 & 64 \\
Warmup ratio & 0.375 & 0.2 & 0.0375 & 0.003 & 0.1 \\
Training epochs & 40 & 15 & 30 & 50 & 30 \\
Drop path & 0.4 & 0.4 & 0.5 & 0.6 & 0.4 \\
Image resolution & 256$^2$ & 512$^2$ & 896$^2$ & 1280$^2$ & 256$^2$ \\
\end{tabular}
\caption{\textbf{Fine-tuning setting for vision tasks.}}
\label{tb:v_task_config}
\end{table*}

\subsection{Audio-(language) Tasks}
\label{app:audio_details}

We describe the implementation details of audio-text retrieval, audio classification, and audio question answering here. All detailed hyperparameters are listed in Table~\ref{tb:audio_config}.

\begin{table*}[h]
\normaltablestyle{8pt}{1.1}
\begin{tabular}{y{96}|cccc}
Config & AudioCaps \& Clotho & FSD50K & VGGSound & AQA \\
\shline
Optimizer & \multicolumn{4}{c}{AdamW} \\
Optimizer momentum & \multicolumn{4}{c}{$\beta_1, \beta_2{=}0.9, 0.999$} \\
Weight decay & \multicolumn{4}{c}{0.05} \\
Gradient clip & \multicolumn{4}{c}{0.0} \\
Warmup ratio & \multicolumn{4}{c}{0.1} \\
Learning rate schedule & \multicolumn{4}{c}{cosine decay} \\
Numerical precision & \multicolumn{4}{c}{$\mathtt{bf16}$} \\
Peak learning rate & 1.5e-4 & 1e-4 & 8e-5 & 7e-5 \\
Layer-wise lr decay & 0.95 & 0.9 & 0.95 & 0.9  \\
Batch size & 384 & 128 & 512 & 128 \\
Training epochs & 10 & 10 & 10 & 10 \\
Drop path & 0.9 & 0.5 & 0.6 & 0.5 \\
Max duration & 20s & 15s & 15s & 15s \\
\end{tabular}
\caption{\textbf{Fine-tuning setting for audio(-language) tasks.}}
\label{tb:audio_config}
\end{table*}

\paragraph{Audio-Text Retrieval}
We evaluate \onepeace on AudioCaps~\cite{audiocaps} and Clotho~\cite{clotho} datasets.
To get better results, we merge the training set of AudioCaps~\cite{audiocaps}, Clotho~\cite{clotho}, and MACS as the fine-tuning dataset.
Similar to image-text retrieval, we use A-Branch and L-Branch to extract the features of audio clips and texts respectively, and then calculate the cosine similarity between these features.
The recall@k is employed as the evaluation metric.

\paragraph{Audio Classification}
We conduct experiments on three datasets: ESC-50~\cite{esc50}, FSD50K~\cite{fsd50k}, and VGGSound~\cite{vggsound}. 
ESC-50 is an environmental sound dataset that contains $2000$ environmental audio recordings and $50$ labels. 
We directly use the pretrained \onepeace model to perform zero-shot audio classification on ESC-50. Specifically, we use A-Branch to extract audio embeddings from the audio clips and use L-Branch to extract text embeddings from the label names. Then we determine the labels of the audio clips by calculating the similarity between the embeddings.
For FSD50K and VGGSound, we input the original audio into the A-Branch and utilize multi-head attention pooling (MAP) to aggregate the features. 
FSD50K is a multi-label sound event dataset, for which we use BCELoss as the loss function and report the mean average precision on the test set.
VGGSound is an audio-visual dataset, where each sample includes a video with audio.
We extract the audio clips from the videos and excluded the visual information, using cross entropy as the loss function and reporting accuracy on the test set.

\paragraph{Audio Question Answering}
We conduct experiments on the AVQA dataset~\cite{avqa}. 
Each sample in this dataset consists of a video, a question, and four candidate answers. 
To perform the audio question answering task, we extract audio clips from the videos and excluded the visual information. 
During training, we concatenate each answer with the audio and question, and extracted the features through AL-Branch. 
We then minimize the pairwise hinge loss between the positive features and negative features.

\subsection{Vision-language tasks}
\label{app:vision_language_details}
Here we describe the implementation details of different vision-language tasks, including image-text retrieval~\cite{flickr,mscoco}, visual grounding~\cite{refcoco,refcocog}, visual question answering~\cite{vqa}, and visual reasoning~\cite{nlvr2}.
All detailed hyperparameters are listed in Table~\ref{tb:vision_language_config}.

\begin{table*}[h]
\normaltablestyle{8pt}{1.2}
\begin{tabular}{y{96}|cccccc}
Config & MSCOCO & Flickr30K & RefCOCO/$+$/g & VQA & NLVR2 \\
\shline
Optimizer & \multicolumn{5}{c}{AdamW} \\
Optimizer momentum & \multicolumn{5}{c}{$\beta_1, \beta_2{=}0.9, 0.999$} \\
Weight decay & \multicolumn{5}{c}{0.05} \\
Gradient clip & \multicolumn{5}{c}{0.0} \\
Warmup ratio & \multicolumn{5}{c}{0.1} \\
Learning rate schedule & \multicolumn{5}{c}{cosine decay} \\
Numerical precision & \multicolumn{5}{c}{$\mathtt{bf16}$} \\
Peak learning rate & 8e-5 & 7e-5 & 1.5e-4 & 3e-4 & 1e-4 \\
Layer-wise lr decay & 0.9 & 0.9 & 0.9 & 0.85 & 0.9 \\
Batch size & 3072 & 3072 & 256 & 512 & 128 \\
Training epochs & 15 & 20 & 30 & 10 & 25 \\
Drop path & 0.5 & 0.4 & 0.4 & 0.5 & 0.4 \\
Image resolution & 432 & 432 & 384 & 768 & 256 \\
\end{tabular}
\caption{\textbf{Fine-tuning setting for vision-language tasks.}}
\label{tb:vision_language_config}
\end{table*}

\paragraph{Image-Text Retrieval}
We evaluate \onepeace on MSCOCO~\cite{mscoco} and Flickr30K~\cite{flickr} datasets, and report the results on the widely used Karpathy test split~\cite{karpathy}.
We use V-Branch and L-Branch to extract the features of images and texts respectively, and then calculate the cosine similarity between these features.
The recall@k is employed as the evaluation metric.

\paragraph{Visual Grounding}
This task requires the model to locate an image region based on a text description.
We conduct experiments on RefCOCO, RefCOCO+, and RefCOCOg datasets~\cite{refcoco,refcocog}.
The image and text are fed to the VL-Branch simultaneously, then we use multi-head attention pooling (MAP)~\cite{set_transformer} to aggregate the features from all image patches.
The pooled output is used to predict the continuous corner coordinates of the bounding box $\left(x_1,y_1,x_2,y_2 \right)$, where $x_1$ and $y_1$ denotes the normalized top left coordinates, $x_2$ and $y_2$ denotes the normalized bottom right coordinates.
We report the standard metric Acc@0.5 on the validation and test sets.

\paragraph{Visual Question Answering}
This task requires the model to answer the question based on an image. We perform experiments on the VQAv2 dataset~\cite{vqav2}.
Following previous works~\cite{ofa,albef,blip,x-vlm}, we use the training and validation set of VQAv2 for training, including additional question-answer pairs from Visual Genome~\cite{vg}.
The image and question are fed to the VL-Branch simultaneously, then we use MAP to aggregate the features from all text tokens.
The pooled output is fed into a classifier to predict the answer from the 3,129 most frequent answers.
We report the final score on the test-dev and test-std sets.


\paragraph{Visual Reasoning}
Given a text and a pair of images, this task requires the model to distinguish whether the text truly describes the images. We conduct experiments on the NLVR2 dataset~\cite{nlvr2}.
Following the common practice, We treat each sample as two image-text pairs, each containing a text and one image. Then we input these pairs into VL-branch respectively.
The final pooled outputs are concatenated together and fed to a classifier to predict the label. 
We report accuracy on the dev and test-P sets.

\section{Effects of Pretrained Audio Feature Extractor}
\label{app:audio_feature_extractor}

We conduct a systematic analysis of the impact of the pretrained audio feature extractor.
We find that although the parameters of the feature extractor are only $4.6$M, accounting for only about $1\%$ of the total parameters, it has a significant impact on the model performance. 
As shown in Table~\ref{tb:audio_ablation}, the feature extractor with random initialization only achieves $85.3$ accuracy on the ESC-50 dataset, while using pretrained feature extractors results in better performance. 
Notably, using the WavLM feature extractor can lead to the largest improvement (+6.2).
 We attribute this to the fact that WavLM is trained on a more diverse audio dataset compared to Hubert and Wav2Vec 2.0, making its feature extractor more suitable for environmental sound tasks.

 \begin{table*}[h]
\centering
\normaltablestyle{6pt}{1.3}
\begin{tabular}{c|cccc}
  Feature Extractor & Random Init. & Init. with Hubert~\cite{hubert} & Init. with Wav2Vec 2.0~\cite{wav2vec2} & Init. with WavLM~\cite{wavlm}
  \\
  \shline
  ESC-50 Acc. & 85.8 &89.6 (+3.8)  & 90.0 (+4.2) & 91.8 (+6.0)
  \\
\end{tabular}
\caption{\textbf{Ablation studies of pretrained audio feature extractors.} We report zero-shot accuracy on the ESC-50 dataset.}
\label{tb:audio_ablation}
\end{table*}

\section{Evaluate \onepeace on \textit{One Piece}}
\label{app:one_piece}
\begin{figure*}[t]
    \centering
    \includegraphics[width=\linewidth]{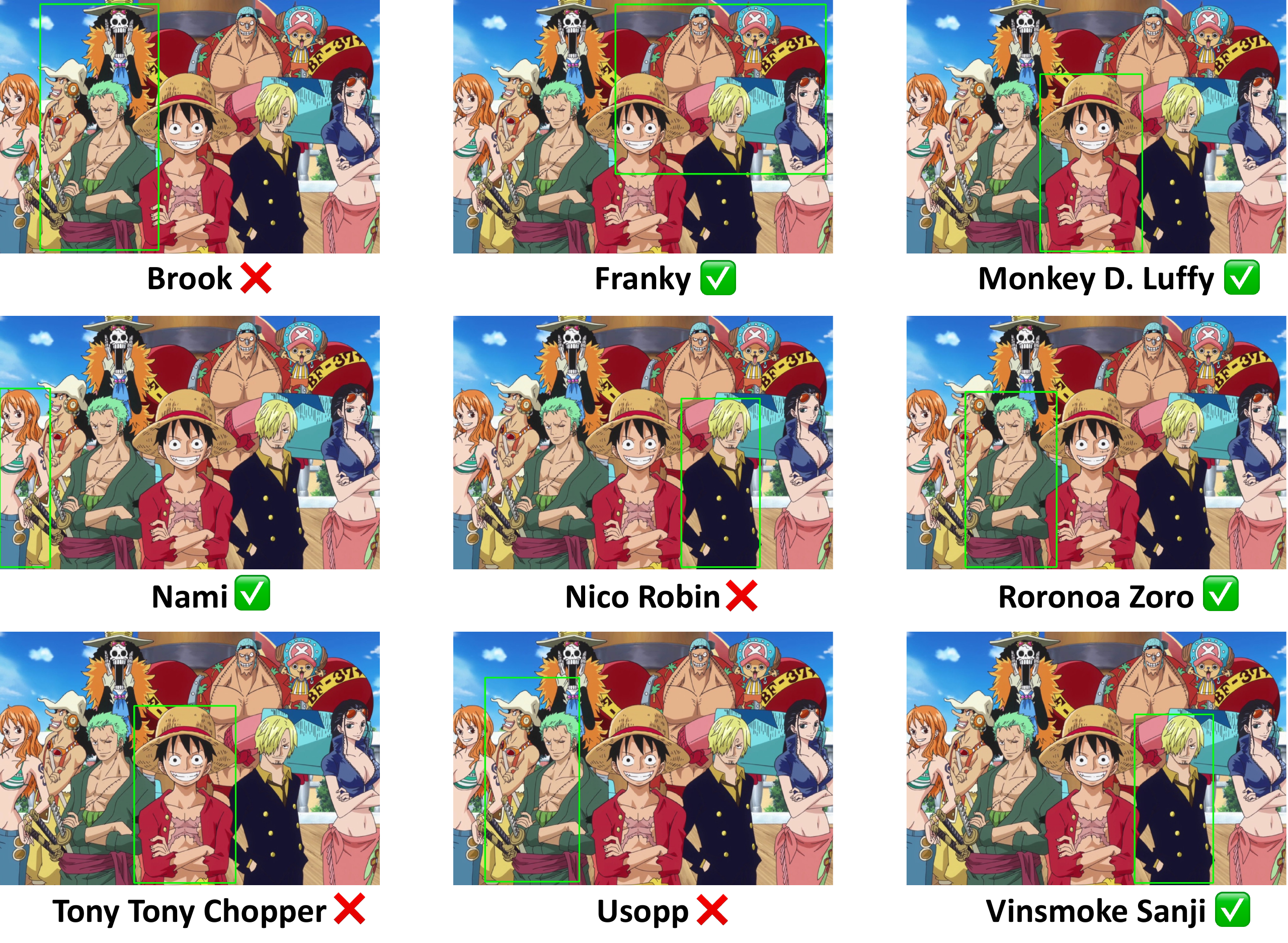}
    \caption{Visualization of \onepeace locating different characters of \textit{One Piece}. Given the names of 9 members of the Straw Hat Pirates, \onepeace correctly located 5 of them from the picture.}
    \label{fig:onepiece_grounding_visual}
\end{figure*}

We further test the visual grounding ability of \onepeace by using a more complex anime picture, \textit{One Piece}. 
The model is fine-tuned on the RefCOCOg dataset.
As shown in Figure~\ref{fig:onepiece_grounding_visual}, we ask \onepeace to locate the characters based on their names. Although \onepeace hasn't seen any anime pictures in the RefCOCOg dataset, it still achieves a recognition accuracy of 56.6\%.

\label{app:visual_grounding_onepiece}

\section{More Examples of Emergent Zero-shot Retrieval}
\label{app:emergent_zero_shot}

In this section, we provide more examples to demonstrate the emergent zero-shot abilities of \onepeace, including audio-to-image, audio+image-to-image, and audio+text-to-image retrieval.
The audios are selected from ESC-50~\cite{esc50}, and the images are retrieve from ImageNet-1K~\cite{imagenet} and MSCOCO~\cite{mscoco}.
By reading this section, we hope that readers can better perceive \onepeace.

\begin{figure*}[t]
    \centering
    \includegraphics[width=\linewidth]{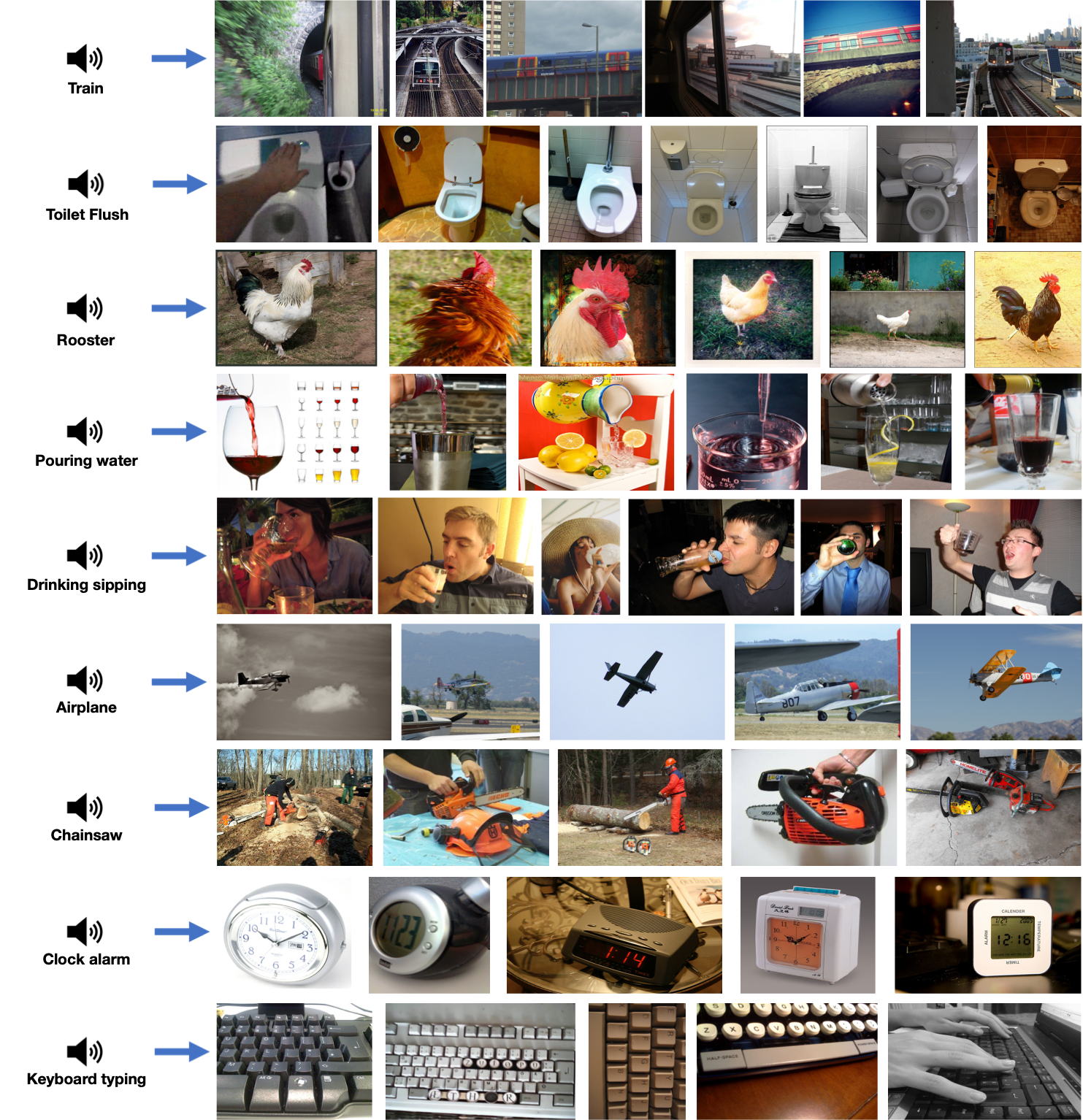}
    \caption{\textbf{Audio-to-image retrieval.}}
    \label{fig:audio2img}
\end{figure*}

\begin{figure*}[t]
    \centering
    \includegraphics[width=\linewidth]{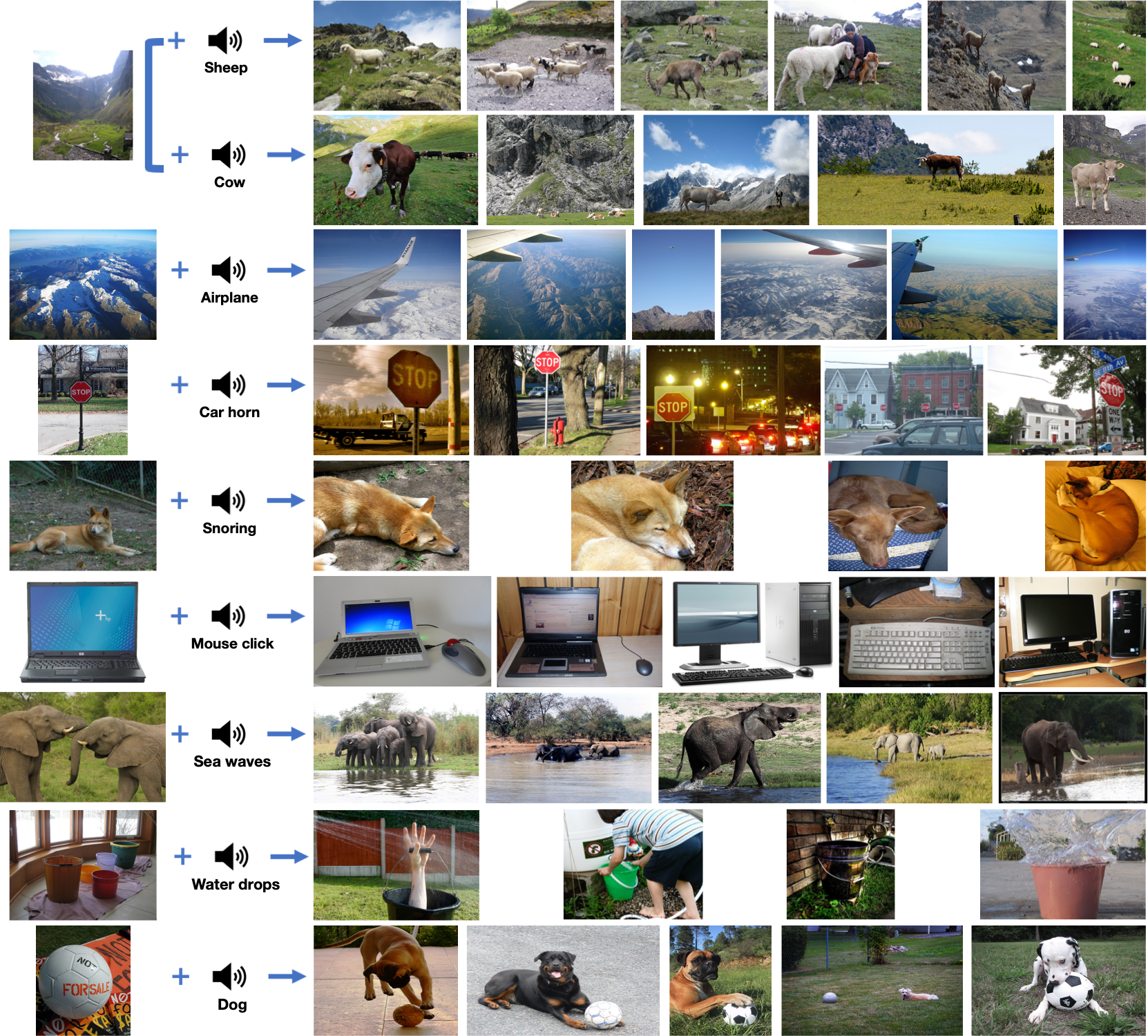}
    \caption{\textbf{Audio+image-to-image retrieval.}}
    \label{fig:audioimg2img}
\end{figure*}

\begin{figure*}[t]
    \centering
    \includegraphics[width=\linewidth]{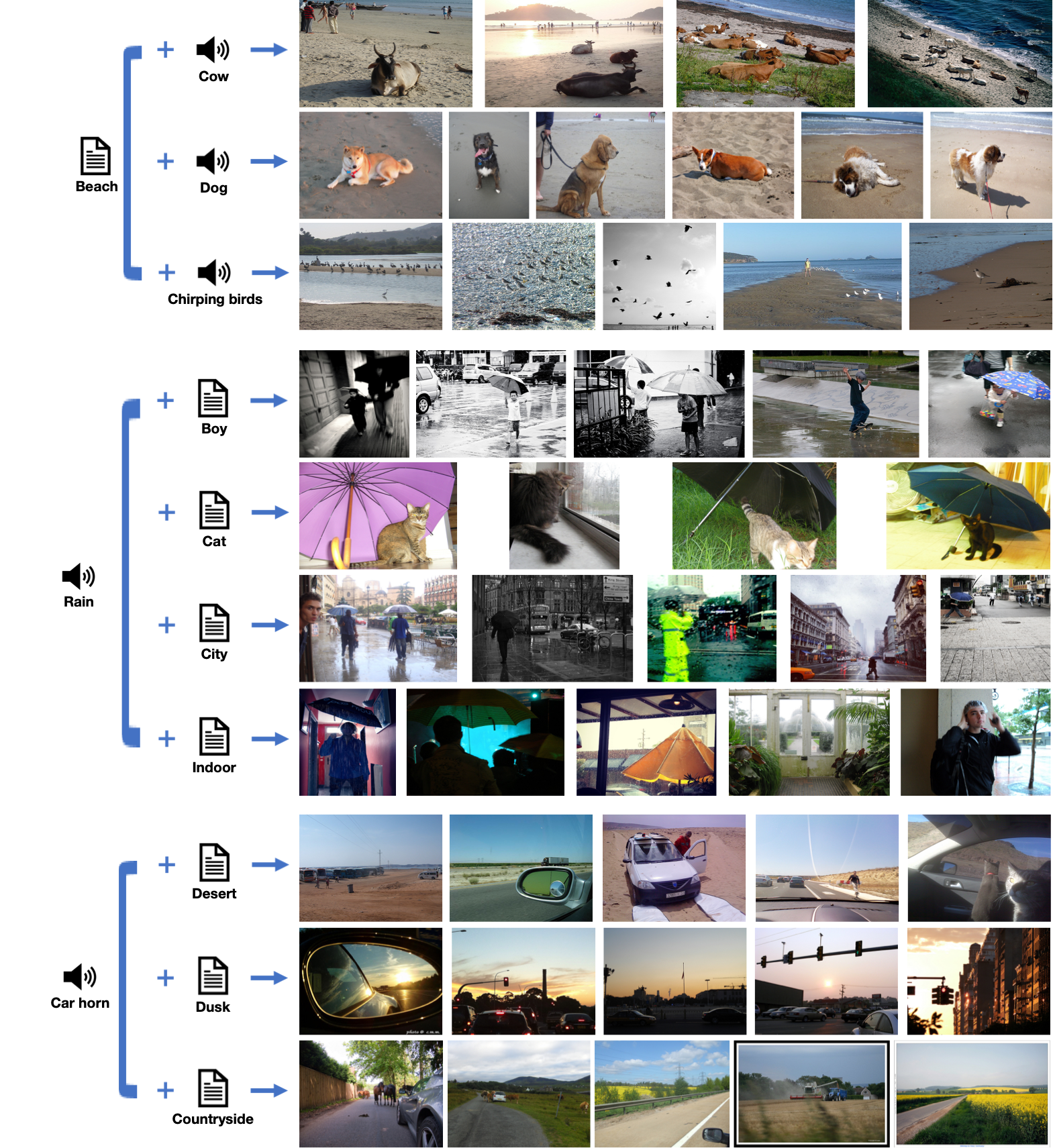}
    \caption{\textbf{Audio+text-to-image retrieval.}}
    \label{fig:audiotext2img}
\end{figure*}

%% file: table/audio_dataset.tex

\begin{table*}[h]
\centering
\normaltablestyle{8pt}{1.2}
\begin{tabular}{l|ccc}
  Dataset
  & Number of Samples
  & Duration
  & T5 Augmentation
  \\
  \shline
  Epidemic Sound & 75618 & 220.40 hrs & Yes
  \\ 
  AudioCaps~\cite{audiocaps} & 49494 & 135.56 hrs & No
  \\
  AudioSet~\cite{audioset} & 1910918 & 5263.23 hrs & Yes
  \\
  AudioStock & 9552 & 40.31 hrs & No
  \\
  Clotho~\cite{clotho} & 3839 & 23.99 hrs & No
  \\ 
  FreeSound~\cite{freesound} & 363618 & 2162.10 hrs & No
  \\
  MACS & 3537 & 9.85 hrs & No
  \\ 
  SoundDescs~\cite{sounddescs} & 10677 & 331.67 hrs & No
  \\
  WavText5K~\cite{wavtext5k} & 2248 & 17.23 hrs & No
  \\
\end{tabular}
\caption{\textbf{Statistics on the environmental sound datasets.} All datasets are publicly available.}
\label{tb:audio_dataset}
\end{table*}